\documentclass[12pt]{article}
\usepackage{arxiv}

\usepackage[utf8]{inputenc} %
\usepackage[T1]{fontenc}    %
\usepackage{booktabs}       %
\usepackage{amsfonts}       %
\usepackage{nicefrac}       %
\usepackage{microtype}      %

\usepackage{url}
\usepackage{xcolor}         %
\usepackage[inline]{enumitem} %

\usepackage[numbers,sort&compress]{natbib}
\usepackage{doi}
\usepackage{algorithm,algpseudocode}
\usepackage{amsmath}
\usepackage{amssymb}
\usepackage{mathtools}
\usepackage{amsthm}
\usepackage{multirow, array, threeparttable}
\usepackage[font=normalsize,skip=4pt]{caption}
\usepackage{subcaption}

\usepackage{graphicx}   %
\hypersetup{
    colorlinks,
    linkcolor={blue!60!black},
    citecolor={red!60!black},
    urlcolor={blue!80!black}
}
\usepackage[capitalize,nameinlink]{cleveref}
\DeclareMathOperator{\diag}{diag}

\makeatletter
\renewcommand{\paragraph}{%
  \@startsection{paragraph}{4}{\z@}%
  {0.25\baselineskip}  %
  {-0.25em}            %
  {\normalfont\normalsize\bfseries}%
}
\makeatother

\usepackage{enumitem}

\newtheorem{theorem}{Theorem}
\newtheorem{proposition}{Proposition}

\algnewcommand\algorithmicinput{\textbf{Input:}}
\algnewcommand\algorithmicoutput{\textbf{Output:}}
\algnewcommand\algorithmicnote{\textbf{Note:}}
\algnewcommand\Input{\item[\algorithmicinput]}%
\algnewcommand\Output{\item[\algorithmicoutput]}%
\algnewcommand\Note{\item[\algorithmicnote]}%

\newcommand{\bigO}{\mathcal{O}}

\title{Optimizing Posterior Samples for Bayesian Optimization via Rootfinding}

\author{{\hspace{1mm}Taiwo Adebiyi} \\
	University of Houston\\
	\texttt{taadebiyi2@uh.edu} \\
        \And
{\hspace{1mm}Bach Do}\\
	University of Houston\\
	\texttt{bdo3@uh.edu} \\
        \And
 {\hspace{1mm}Ruda Zhang} \\
	University of Houston\\
	\texttt{rudaz@uh.edu} \\
}

\newcommand{\matern}{{Mat\'ern }} %

\newcommand{\diff}{\,\mathrm{d}} %
\newcommand{\dset}{\{1, \cdots, d\}} %
\newcommand{\ximin}{\underline{x}_i} %
\newcommand{\ximax}{\overline{x}_i} %
\newcommand{\xjmin}{\underline{x}_j} %
\newcommand{\xjmax}{\overline{x}_j} %
\newcommand{\xdata}{\mathbf{x}^i} %
\newcommand{\ftrue}{f_{\text{true}}} %
\newcommand{\iid}{\overset{\text{iid}}{\sim}} %
\newcommand{\eqdist}{\overset{\text{d}}{=}} %
\newcommand{\randprior}{\omega} %
\newcommand{\randpost}{\widetilde{\omega}} %
\newcommand{\fprior}{f_{\randprior}} %
\newcommand{\fpost}{\widetilde{f}_{\randpost}} %
\newcommand{\fadjust}{b} %
\newcommand{\meanpost}{\widetilde{\mu}} %
\newcommand{\foptpost}{\widetilde{f}_{\randpost}^\star} %
\newcommand{\xoptpost}{\widetilde{\mathbf{x}}_{\randpost}^\star} %
\newcommand{\fopt}{\widetilde{f}^\star} %
\newcommand{\xopt}{\widetilde{\mathbf{x}}^\star} %
\newcommand{\xlocminprior}{\breve{\mathbf{x}}_{\randprior}} %
\newcommand{\Xlocminprior}{\breve{X}_{\randprior}} %
\newcommand{\xlocminpost}{\breve{\widetilde{\mathbf{x}}}_{\randpost}} %
\newcommand{\Xlocminpost}{\breve{\widetilde{X}}_{\randpost}} %
\newcommand{\Xlocmaxprior}{\widehat{X}_{\randprior}} %
\newcommand{\Xneglocminprior}{\breve{X}_{\randprior}^{-}} %
\newcommand{\Xposlocminprior}{\breve{X}_{\randprior}^{+}} %
\newcommand{\Xlocmin}{\breve{X}} %
\newcommand{\Xlocmax}{\widehat{X}} %
\newcommand{\Xneglocmin}{\breve{X}^{-}} %
\newcommand{\Xposlocmin}{\breve{X}^{+}} %
\newcommand{\Xexplore}{S_{\text{e}}} %
\newcommand{\Xexploit}{S_{\text{x}}} %
\newcommand{\xicritical}{\mathring{\Xi}_i} %
\newcommand{\xicandidate}{\Xi_i} %
\newcommand{\xicandidatemono}{\Xi_i^{(0)}} %
\newcommand{\xicandidatemixed}{\Xi_i^{(1)}} %
\newcommand{\xicandidatepos}{\Xi_i^{+}} %
\newcommand{\xicandidateneg}{\Xi_i^{-}} %
\newcommand{\xicandidateposmono}{\Xi_i^{+(0)}} %
\newcommand{\xicandidateposmixed}{\Xi_i^{+(1)}} %
\newcommand{\xicandidatenegmono}{\Xi_i^{-(0)}} %
\newcommand{\xicandidatenegmixed}{\Xi_i^{-(1)}} %
\newcommand{\Xmono}{\Xi^{(0)}} %
\newcommand{\Xmixed}{\Xi^{(1)}} %
\newcommand{\Xcandidate}{S_{\text{o}}} %
\newcommand{\Xprecandidate}{\breve{S}^{-}} %
\newcommand{\Xmixedmag}{S^{(1)}} %
\newcommand{\nexplore}{n_{\text{e}}} %
\newcommand{\nexploit}{n_{\text{x}}} %
\newcommand{\ncandidate}{n_{\text{o}}} %
\DeclareMathOperator*{\argmin}{arg\,min} %
\DeclareMathOperator*{\argmaxk}{arg\,maxk} %
\DeclareMathOperator*{\argmink}{arg\,mink} %
\DeclareMathOperator{\sign}{sign} %
\DeclareMathOperator*{\interior}{int} %

\date{}

\renewcommand{\headeright}{ }
\renewcommand{\shorttitle}{TS-roots}

\begin{document}
\maketitle

\begin{abstract}
Bayesian optimization devolves the global optimization of a costly objective function
to the global optimization of a sequence of acquisition functions.
This inner-loop optimization can be catastrophically difficult if it involves posterior sample paths,
especially in higher dimensions.
We introduce an efficient global optimization strategy for posterior samples
based on global rootfinding.
It provides gradient-based optimizers with two sets of
judiciously selected starting points, designed to combine exploration and exploitation.
The number of starting points can be kept small without sacrificing optimization quality.
Remarkably, even with just one point from each set, the global optimum is discovered most of the time.
The algorithm scales practically linearly to high dimensions, breaking the curse of dimensionality.
For Gaussian process Thompson sampling (GP-TS),
we demonstrate remarkable improvement in both inner- and outer-loop optimization,
surprisingly outperforming alternatives like EI and GP-UCB in most cases.
Our approach also improves the performance of other posterior sample-based acquisition functions,
such as variants of entropy search.
Furthermore, we propose a sample-average formulation of GP-TS,
which has a parameter to explicitly control exploitation
and can be computed at the cost of one posterior sample.
Our implementation is available at \url{https://github.com/UQUH/TSRoots}.
\end{abstract}

\section{Introduction}
\label{introduction}

Bayesian optimization (BO) is a highly successful approach to
the global optimization of expensive-to-evaluate black-box functions,
with applications ranging from hyper-parameter training of machine learning models
to scientific discovery and engineering design \citep{Jones1998,Snoek2012,Frazier2018,Garnett2023}.
Many BO strategies are also backed by strong theoretical guarantees
on their convergence to the global optimum \citep{Srinivas2010, Bull2011, Russo2014, Chowdhury2017}.

Consider the global optimization problem $\min_{\mathbf{x} \in \mathcal{X}} f(\mathbf{x})$
where $\mathbf{x} \in \mathcal{X} \subset \mathbb{R}^d$ represents the vector of input variables and $f(\mathbf{x}) \in \mathbb{R}$ the objective function which can be evaluated at a significant cost,
subject to observation noise.
At its core, BO is a sequential optimization algorithm 
that uses a probabilistic model of the objective function to guide its evaluation decisions.
Starting with a prior probabilistic model and some initial data,
BO derives an acquisition function $\alpha(\mathbf{x})$ from the posterior model,
which is much easier to evaluate than the objective function
and often has easy-to-evaluate derivatives.
The acquisition function is then optimized globally, using off-the-shelf optimizers,
to provide a location to evaluate the objective function.
This process is iterated until some predefined stopping criteria are met.

Effectively there are two nested iterations in BO:
the outer-loop seeks to optimize the objective function $f(\mathbf{x})$,
and the inner-loop seeks to optimize the acquisition function $\alpha(\mathbf{x})$ at each BO iteration.
The premise of BO is that the inner-loop optimization can be solved accurately and efficiently,
so that the outer-loop optimization proceeds informatively with a negligible added cost.
In fact, the convergence guarantees of many BO strategies
assume \textit{exact} global optimization of the acquisition function.
However, the efficient and accurate global optimization of acquisition functions
is less trivial than it is often assumed to be \citep{Wilson2018}.

Acquisition functions are, in general, highly non-convex and have many local optima.
In addition, many common acquisition functions
are mostly flat surfaces with a few peaks \citep{Rana2017}, %
which take up an overwhelmingly large portion of the domain as the input dimension grows.
This creates a significant challenge for generic global optimization methods.

Some acquisition functions involve sample functions from the posterior model,
which need to be optimized globally.
Gaussian process Thompson sampling (GP-TS) \citep{Chowdhury2017} uses posterior sample paths directly as random acquisition functions.
In many information-theoretic acquisition functions
such as entropy search (ES) \citep{Hennig2012}, predictive entropy search (PES) \citep{HernandezLobato2014},
max-value entropy search (MES) \citep{WangZ2017}, and joint entropy search (JES) \citep{Tu2022,Hvarfner2022},
multiple posterior samples are drawn and optimized to find their global optimum location and/or value.
Such acquisition functions are celebrated for their nice properties in BO:
TS has strong theoretical guarantees \citep{Russo2014, Russo2016}
and can be scaled to high dimensions \citep{Mutny2018};
information-theoretic acquisition functions are grounded
in principles for optimal experimental design \citep{MacKay2003};
and both types can be easily parallelized in synchronous batches
\citep{Shah2015, HernandezLobato2017, Kandasamy2018}.
However, posterior sample paths are much more complex than other acquisition functions,
as they fluctuate throughout the design space, and are less smooth than the posterior mean and marginal variance.
The latter are the basis of many acquisition functions,
such as expected improvement (EI) \citep{Jones1998}, probability of improvement (PI) \citep{Kushner1964},
and upper confidence bound (GP-UCB) \citep{Srinivas2010}.
As a consequence, posterior sample paths have many more local optima,
and the number scales exponentially with the input dimension.

While there is a rich literature on prior probabilistic models and acquisition functions for BO,
global optimization algorithms for acquisition functions have received little attention.
One class of global optimization methods is derivative-free,
such as the dividing rectangles (DIRECT) algorithm \citep{Jones1993},
covariance matrix adaptation evolution strategy (CMA-ES) algorithms \citep{Hansen2003},
and genetic algorithms \citep{Mitchell1998}.
Gradient-based multistart optimization, on the other hand,
is often seen as the best practice to reduce the risk of getting trapped in local minima \citep{Kim2021},
and enjoys the efficiency of being embarrassingly parallelizable.
For posterior samples, their global optimization may use
joint sampling on a finite set of points \citep{Kandasamy2018},
or approximate sampling of function realizations followed by gradient-based optimization
\citep{HernandezLobato2014, Mutny2018}.
The selection of starting points is crucial for the success of gradient-based multistart optimization.
This selection can be deterministic (e.g., grid search), random \citep{Bergstra2012,Balandat2020}, or adaptive \citep{Feo1995}.

In this paper, we propose an adaptive strategy for selecting starting points for
gradient-based multistart optimization of posterior samples.
This algorithm builds on the decomposition of posterior samples by pathwise conditioning, %
taps into robust software in univariate function computation based on univariate global rootfinding,
and exploits the separability of multivariate GP priors.
Our key contributions include:
\begin{itemize}[leftmargin=*,topsep=0pt,itemsep=-1ex,partopsep=1ex,parsep=1ex]
	\item A novel strategy for the global optimization of posterior sample paths.
	The starting points are dependent on the posterior sample,
	so that each is close to a local optimum that is a candidate for the global optimum.
	The selection algorithm scales linearly to high dimensions.
	\item We give empirical results across a diverse set of problems
	with input dimensions ranging from 2 to 16,
	establishing the effectiveness of our optimization strategy.
	Although our algorithm is proposed for the inner-loop optimization of posterior samples,
	perhaps surprisingly, we see significant improvement in outer-loop optimization performance,
	which often allows acquisition functions based on posterior samples to converge faster
	than other common acquisition functions.
	\item A new acquisition function via the posterior sample average that explicitly controls the exploration--exploitation balance \citep{Chapelle2011},
	and can be generated at the same cost as one posterior sample.
\end{itemize}

\section{General Background}
\label{sec:background}

\paragraph{Gaussian Processes.}

Consider an unknown function $\ftrue: \mathcal{X} \mapsto \mathbb{R}$, where domain $\mathcal{X} \subset \mathbb{R}^d$.
We can collect noisy observations of the function through the model $y^i = \ftrue(\mathbf{x}^i) + \varepsilon^i$,
$i \in \{1, \cdots, n\}$, with $\boldsymbol{\varepsilon} \sim \mathcal{N}_n(0, \boldsymbol{\Sigma})$.
To model the function $\ftrue$, we use a Gaussian process (GP) as the prior probabilistic model: $f \sim \pi \in \mathcal{GP}$.
A GP is a random function $f$ such that for any finite set of points $X = \{\mathbf{x}^i\}_{i=1}^n$, $n \in \mathbb{N}$,
the values $\mathbf{f}_n = (f(\mathbf{x}^i))_{i=1}^n$ have a multivariate Gaussian distribution $\mathcal{N}_n(\boldsymbol{\mu}_n, \mathbf{K}_{n,n})$,
with mean $\boldsymbol{\mu}_n = (\mu(\mathbf{x}^i))_{i=1}^n$
and covariance $\mathbf{K}_{n,n} = [\kappa(\mathbf{x}^i, \mathbf{x}^j)]_{i \in n}^{j \in n}$. Here, $\mu(\mathbf{x})$ is the mean function and $\kappa(\mathbf{x}, \mathbf{x}')$ is the covariance function.

\paragraph{Decoupled Representation of GP Posteriors.}

Given a dataset $\mathcal{D} = \{(\mathbf{x}^i, y^i)\}_{i=1}^n$,
the posterior model $f | \mathcal{D}$ is also a GP.
Samples from the posterior have a decoupled representation called pathwise conditioning,
originally proposed in \citep{Wilson2020, Wilson2021}:
\begin{equation}\label{eq:posterior-decoupled}
    (f|\mathcal{D})(\mathbf{x}) \eqdist f(\mathbf{x})
    + \boldsymbol{\kappa}_{\cdot,n}(\mathbf{x}) (\mathbf{K}_{n,n} + \boldsymbol{\Sigma})^{-1}
    (\mathbf{y} - \mathbf{f}_n - \boldsymbol{\varepsilon}),
    \quad f \sim \pi,
    \; \boldsymbol{\varepsilon} \sim \mathcal{N}_n(0, \boldsymbol{\Sigma}),
\end{equation}
where $f(\mathbf{x})$ is a sample path from the GP prior,
$\boldsymbol{\kappa}_{\cdot,n}(\mathbf{x}) = (\kappa(\mathbf{x}, \mathbf{x}^i))_{i=1}^n$ is the canonical basis,
$\mathbf{f}_n = (f(\mathbf{x}^i))_{i=1}^n$, and $\boldsymbol{\varepsilon}$ is a sample of the noise.
We may interpret $\mathbf{f}_n + \boldsymbol{\varepsilon}$ as a sample from the prior distribution of the observations
$\mathbf{y} = (y^i)_{i=1}^n$.
This representation has its roots in Matheron's update rule that transforms a joint distribution of Gaussian variables into a conditional one (see e.g., \cite{Hoffman1991}).
This formula is exact, in that $\eqdist$ denotes equality in distribution,
and it preserves the differentiability of the prior sample.
It is also computationally efficient for posterior sampling:
the cost is independent of the input dimension $d$,
linear in the data size $n$ at evaluation time,
and the weight vector for $\boldsymbol{\kappa}_{\cdot,n}(\mathbf{x})$
can be solved accurately using an iterative algorithm that scales linearly with $n$ \citep{LinJ2023}.

\section{Global Optimization of Posterior Samples}
\label{sec:methods}

In this section, we introduce an efficient algorithm for the global optimization of posterior sample paths.
For this, we exploit the separability of prior samples and useful properties of posterior samples
to judiciously select a set of starting points for gradient-based multistart optimizers.

\paragraph{Assumptions.}

Following \Cref{sec:background}, we impose a few common assumptions throughout this paper:
(1) the domain is a hypercube: $\mathcal{X} = \prod_{i=1}^d [\ximin, \ximax]$;
(2) prior covariance is separable: $\kappa(\mathbf{x}, \mathbf{x}') = \prod_{i=1}^d \kappa_i(x_i, x_i')$;
(3) prior samples are continuously differentiable: $f(\mathbf{x}; \randprior) \in C^1$.
Without loss of generality, we also assume that the prior mean $\mu(\mathbf{x}) = 0$:
any non-zero mean function can be subtracted from the data by replacing $\ftrue$ with $\ftrue - \mu$.
While additive and multiplicative compositions of univariate kernels can be used in the prior \citep{Duvenaud2013},
assumption (2) is the most popular choice in BO.
It is possible to extend our method to generalized additive models.

\alglanguage{pseudocode}
\begin{algorithm}[t]
	\caption{\texttt{TS-roots}: Optimizing posterior samples via rootfinding}
	\label{alg:TSroots}
	\begin{algorithmic}[1] %
		\Input prior samples $f(\mathbf{x})$, $\boldsymbol{\varepsilon}$;
            prior covariances $\kappa(\mathbf{x}, \mathbf{x}')$, $\boldsymbol{\Sigma}$;
            dataset $\mathcal{D}$; set sizes $\ncandidate, \nexplore, \nexploit$.
		\State $\Xcandidate \gets \texttt{minsort}(f(\mathbf{x}), \ncandidate)$
            \Comment{Smallest $\ncandidate$ local minima of the prior sample (\Cref{alg:minsort})}
		\State $[\widetilde{\mathbf{f}}_{\text{e}}, I_{\text{e}}] \gets \texttt{mink}(\widetilde{f}(\Xcandidate), \nexplore)$
            \Comment{Smallest $\nexplore$ values and indices of the posterior sample in $\Xcandidate$}
		\State $[\widetilde{\mathbf{f}}_{\text{x}}, I_{\text{x}}] \gets \texttt{mink}(\widetilde{f}(X), \nexploit)$
            \Comment{Smallest $\nexploit$ values and indices of the posterior mean in $X$}
            \State $\Xexplore \gets \Xcandidate[I_{\text{e}}, :], \; \Xexploit \gets X[I_{\text{x}}, :]$
            \Comment{Starting points: exploration set and exploitation set}
		\State $[\xopt, \fopt] \gets \texttt{minimize}(\widetilde{f}(\mathbf{x}), S), \; S = \Xexplore \cup \Xexploit$
            \Comment{Gradient-based multistart optimization}
		\Output Thompson sample $\xopt$
            \Comment{Global minimum of the posterior sample}
	\end{algorithmic}
\end{algorithm}

\subsection{TS-roots Algorithm}

We observe that, given the assumptions, a posterior sample in \cref{eq:posterior-decoupled} can be rewritten as:
\begin{equation}\label{eq:posterior-sample}
\widetilde{f}(\mathbf{x}; \randpost) = f(\mathbf{x}; \randprior) + b(\mathbf{x}; \randpost),
\quad f(\mathbf{x}; \randprior) \approx \prod_{i=1}^d f_i(x_i; \randprior_i),
\quad b(\mathbf{x}; \randpost) = \sum_{j=1}^n v_j \kappa(\mathbf{x}, \mathbf{x}^j).
\end{equation}
Here, the prior sample $f(\mathbf{x})$ is approximated as a separable function determined by the random bits $\randprior$ (see \Cref{sub:spectral-sampling}).
Data adjustment $b(\mathbf{x})$ is a sum of the canonical basis with coefficients
$\mathbf{v} = (\mathbf{K}_{n,n} + \boldsymbol{\Sigma})^{-1} (\mathbf{y} - \mathbf{f}_n - \boldsymbol{\varepsilon})$.
Both the data adjustment and the posterior sample are determined by the random bits
$\randpost = (\randprior, \boldsymbol{\varepsilon})$.
In the following, we denote the prior and the posterior samples as $\fprior$ and $\fpost$, respectively.
Our goal is to find the global minimum $(\xoptpost, \foptpost)$ of the posterior sample $\fpost(\mathbf{x})$.

The global minimization of a generic function, in principle, requires finding all its local minima and then selecting the best among them.
However, computationally efficient approaches to this problem are lacking even in low dimensions
and, more pessimistically, the number of local minima grows exponentially as the domain dimension increases.
The core idea of this work is to use the prior sample $\fprior$ as a surrogate of the posterior sample $\fpost$ for global optimization.
Another key is to exploit the separability of the prior sample for
efficient representation and ordering of its local extrema.

We define \textit{TS-roots} as a global optimization algorithm for GP posterior samples
(given the assumptions) via gradient-based multistart optimization.
The starting points include:
(1) a subset $\Xexplore$ of the local minima $\Xlocminprior$ of a corresponding prior sample;
and (2) a subset $\Xexploit$ of the observed locations $X$.
We call $\Xexplore$ the exploration set and $\Xexploit$ the exploitation set.
Specifically, let
\begin{equation} \label{eq:Xcandidate}
    \Xcandidate = \argmink_{\mathbf{x} \in \Xlocminprior} (\fprior(\mathbf{x}), \ncandidate)
\end{equation}
be the $\ncandidate$ smallest local minima of the prior sample.
The set of starting points, $S$, is defined as:
\begin{equation} \label{eq:starting_points}
    S = \Xexplore \cup \Xexploit,
    \quad \Xexplore = \argmink_{\mathbf{x} \in \Xcandidate} (\fpost(\mathbf{x}), \nexplore),
    \quad \Xexploit = \argmink_{\mathbf{x} \in X} (\fpost(\mathbf{x}), \nexploit).
\end{equation}
\Cref{alg:TSroots} outlines the procedure for TS-roots.
Here, $\nexplore$ and $\nexploit$ are imposed to control the cost of gradient-based multistart optimization,
and $\ncandidate$ is set to limit the number of evaluations of $\fpost$ in the determination of $\Xexplore$. 
Considering the cost difference, we can have $\ncandidate \gg \nexplore$.
We observe that $\nexplore$ and $\nexploit$ can be set to small values, and $\ncandidate$ to a moderate value,
without sacrificing the quality of optimization, see \Cref{apd:Minsize}.
The TS-roots algorithm scales linearly in $d$, see \Cref{sub:complexity}.

\begin{figure}[t]
\centering
\includegraphics[width=\textwidth]{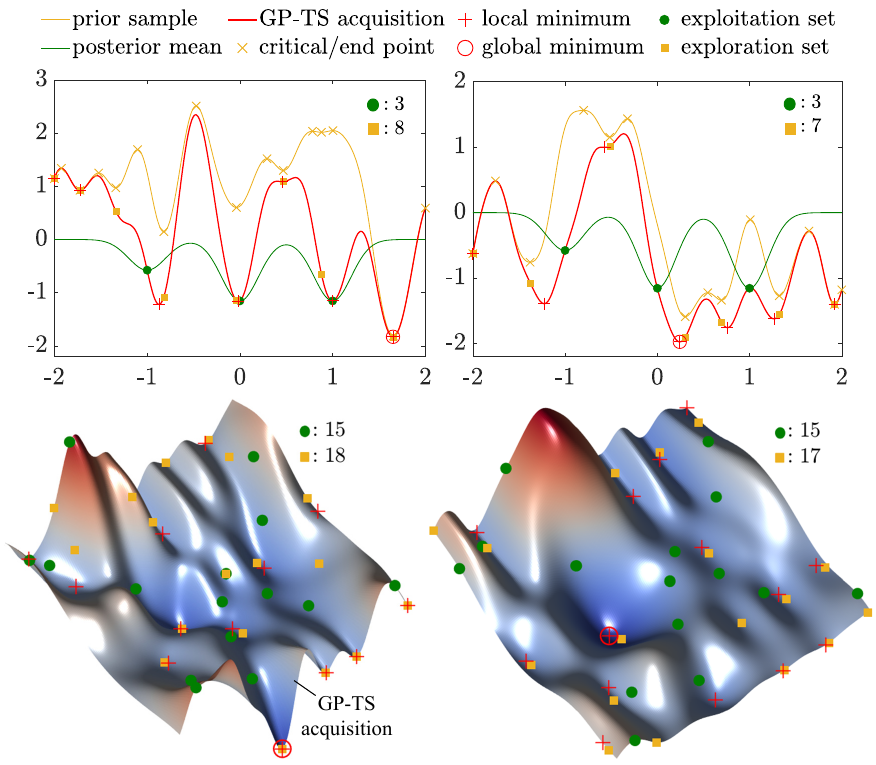}
\caption{Illustrations of exploration and exploitation sets for the global optimization of GP-TS acquisition functions
in one dimension (top row) and two dimensions (bottom row).
\textit{Left column:} When the global minimum $\xoptpost$ of the GP-TS acquisition function lies outside the interpolation region,
it is typically identified by starting the gradient-based multistart optimizer at a local minimum of the prior sample.
\textit{Right column:} When $\xoptpost$
is within the interpolation region,
it can be found by starting the optimizer at either an observed location or a local minimum of the prior sample.} 
\label{fig:ExploreExploitSets}
\end{figure}

\subsection{Relations between the Local Minima of Prior and Posterior Samples}

\Cref{fig:ExploreExploitSets} shows several posterior samples $\fpost$ in one and two dimensions,
each marked with its local minima $\Xlocminpost$ and global minimum $\xoptpost$.
Here the exploration set $\Xexplore$ is the local minima $\Xlocminprior$ of %
$\fprior$,
and the exploitation set $\Xexploit$ is the observed locations $X$.
We make the following observations:
\begin{enumerate}[wide, labelwidth=!, labelindent=0pt, label=(\arabic*),itemsep=1pt]
\item The prior sample $\fprior$ is more complex than the data adjustment $\fadjust$
in the sense that it is less smooth and has more critical points.
The comparison of smoothness can be made rigorous in various ways: for example,
for GPs with a \matern covariance function where the smoothness parameter is finite,
$\fprior$ is almost everywhere one time less differentiable than $\fadjust$
(see e.g., \citet{Garnett2023} Sec. 10.2, \citet{Kanagawa2018}).
\item Item (1) implies that
  when the prior sample $\fprior$ is added to the smoother landscape of $\fadjust$,
  each local minimum $\xlocminprior$ of $\fprior$ will be located near a local minimum $\xlocminpost$ of
  the posterior sample $\fpost$.
  Away from the observed locations $X$, each $\xlocminpost$
  is closely associated with an $\xlocminprior$, with minimal change in location.
  In the vicinity of $X$, an $\xlocminpost$ may have
  both a data point $\xdata$ and an $\xlocminprior$ nearby,
  but because of the smoothness difference of $\fprior$ and $\fadjust$,
  in most cases the one closest to $\xlocminpost$ is an $\xlocminprior$.
\item It is possible that near $X$, sharp changes in $\fprior$
  may require sharp adjustments to the data,
  which may move some $\xlocminprior$ by a significant distance,
  or create new $\xlocminpost$
  that are not near any $\xlocminprior$ or any $\xdata$.
\item Searching from $\xdata$ with good observed values can discover
  good $\xlocminpost$ in the vicinity of $X$,
  which can pick up some local minima not readily discovered by $\Xlocminprior$.
  This is especially true if $\fprior$ is relatively flat near $\xdata$.
\item Since the posterior sample $\fpost$ adapts to the dataset, searches from $\xdata$ will tend to converge to
  a good $\xlocminpost$ among all the local optima near $\xdata$.
  Even if the searches from $X$ cannot discover all the local minima in its vicinity,
  they tend to discover a good subset of them.
  Therefore, (4) can help address the issue in (3), if not fully eliminating it.
  By combining subsets of $\Xlocminprior$ and $X$, 
we can expect that the set of local minima of $\fpost$ discovered with these starting points
includes the global minimum $\xoptpost$ with a high probability with respect to $\randpost$.
\end{enumerate}

\subsection{Representation of Prior Sample Local Minima}
\label{sec:representation_extrema}

For each component function $f_i(x_i; \randprior_i)$ of the prior sample $\fprior(\mathbf{x})$,
define $h_i(\ximin) = f_i'(\ximin)$, $h_i(\ximax) = -f_i'(\ximax)$, and $h_i(x_i) = f_i''(x_i)$ for $x_i \in (\ximin, \ximax)$.
We call a coordinate $\xi_i \in [\ximin, \ximax]$ of \textit{mono type} if $f_i(\xi_i) h_i(\xi_i) > 0$
and call it of \textit{mixed type} if $f_i(\xi_i) h_i(\xi_i) < 0$.
Let $\xicritical = \{\xi_{i,j}\}_{j=1}^{r_i}$ be the set of interior critical points of $f_i$
such that $\xi_{i,j} \in (\ximin, \ximax)$ and $f_i'(\xi_{i,j}) = 0$, $j \in \{1, \cdots, r_i\}$.
Denote $\xi_{i,0} = \ximin$ and $\xi_{i,r_i + 1} = \ximax$.
Partition the set of candidate coordinates $\xicandidate = \{\xi_{i,j}\}_{j=0}^{r_i+1}$
into mono type $\xicandidatemono$ and mixed type $\xicandidatemixed$.
\Cref{prop:representation_extrema} gives a representation of the sets of strong local extrema of the prior sample.
Its proof and the set sizes therein are given in \Cref{apd:proofs}.

\begin{proposition} \label{prop:representation_extrema}
The set of strong local minima of the prior sample $\fprior(\mathbf{x})$ can be written as:
\begin{equation} \label{eq:Xlocminprior}
    \Xlocminprior = \Xneglocminprior \sqcup \Xposlocminprior, \quad
    \Xneglocminprior = \{ \boldsymbol{\xi} \in \Xmixed : \fprior(\boldsymbol{\xi}) < 0 \}, \quad
    \Xposlocminprior = \{ \boldsymbol{\xi} \in \Xmono : \fprior(\boldsymbol{\xi}) > 0 \},
\end{equation}
where tensor grids $\Xi^{(j)} = \prod_{i=1}^d \xicandidate^{(j)}$, $j \in \{0, 1\}$.
The set $\Xlocmaxprior$ of strong local maxima of $\fprior(\mathbf{x})$ has an analogous representation,
and satisfies $\Xlocmaxprior \sqcup \Xlocminprior = \Xmono \sqcup \Xmixed$, where $\sqcup$ is the disjoint union.
\end{proposition}

\paragraph{Critical Points of Univariate Functions via Global Rootfinding.}
To compute the set $\xicritical$ of all critical points of $f_i$
is to compute all the roots of the derivative $f'_i$ on the interval $[\ximin, \ximax]$.
Since $f'_i$ is continuous, this can be solved robustly and efficiently
by approximating the function with a Chebyshev or Legendre polynomial
and solving a structured eigenvalue problem (see e.g., \citet{Trefethen2019}).
The \texttt{roots} algorithm for global rootfinding based on polynomial approximation
is given as \Cref{alg:roots} in \Cref{apd:algorithms}.

\subsection{Ordering of Prior Sample Local Minima}
\label{sec:ordering_minima}

While the size of $\Xlocminprior$ grows exponentially in domain dimension $d$,
its representation in \cref{eq:Xlocminprior} enables an efficient algorithm
to compute the best subset $\Xcandidate$ (\cref{eq:Xcandidate}) without enumerating its elements.

With \cref{eq:Xlocminprior},
we see that $\Xneglocminprior$ consists of all the local minima of $\fprior$ with negative function values.
Consider the case where $\Xneglocminprior$ has at least $\ncandidate$ elements
so that in the definition of $\Xcandidate$ we can replace $\Xlocminprior$ with $\Xneglocminprior$,
which in turn can be replaced with $\Xmixed$.
As we will show later, the problem of finding the largest elements of $|\fprior(\mathbf{x})|$ within $\Xmixed$
is easier than finding the smallest negative elements of $\fprior(\mathbf{x})$.
Once the former is solved, we can solve the latter simply by removing the positive elements.
Therefore, we convert the problem of \cref{eq:Xcandidate} into three steps:
\begin{enumerate}
    \item $\Xmixedmag = \argmaxk_{\mathbf{x} \in \Xmixed} (|\fprior(\mathbf{x})|, \alpha \ncandidate)$,
    with buffer coefficient $\alpha \ge 1$;
    \item $\Xprecandidate = \{ \mathbf{x} \in \Xmixedmag : \fprior(\mathbf{x}) < 0 \}$,
    so that $\Xprecandidate \subseteq \Xneglocminprior$;
    \item $\Xcandidate = \argmink_{\mathbf{x} \in \Xprecandidate} (\fprior(\mathbf{x}), \ncandidate)$,
    assuming that $|\Xprecandidate| \ge \ncandidate$.
\end{enumerate}
The last two steps are by enumeration and straightforward.
The first step can be solved efficiently using min-heaps,
with a time complexity that scales linearly in $\sum_{i=1}^d |\xicandidatemixed|$ rather than $\prod_{i=1}^d |\xicandidatemixed|$,
see \Cref{apd:ordering}.
The coefficient $\alpha$ is chosen so that $|\Xprecandidate| \ge \ncandidate$.
The case when $|\Xneglocminprior| < \ncandidate < |\Xlocminprior|$ can be handled similarly.
If $\ncandidate \ge |\Xlocminprior|$, no subsetting is needed.
The overall procedure to compute $\Xcandidate$ is given in \Cref{alg:minsort} in \Cref{apd:algorithms}.

\section{Sample-average Posterior Function}
\label{averageTS}
We finally propose a sample-average posterior function that explicitly controls the exploration--exploitation balance
and, notably, can be generated at the cost of generating one posterior sample. Let $\meanpost(\mathbf{x}) = \boldsymbol{\kappa}_{\cdot,n}(\mathbf{x}) (\mathbf{K}_{n,n} + \boldsymbol{\Sigma})^{-1} \mathbf{y}$
be the posterior mean function.
For noiseless observations with $\widetilde{\omega} = \omega$,
we can rewrite the GP posterior function in \cref{eq:posterior-sample} as
$\widetilde{f}_\omega(\mathbf{x}) = \fprior(\mathbf{x}) + \meanpost(\mathbf{x}) + \xi_\omega(\mathbf{x})$,
where $\xi_\omega(\mathbf{x}) = - \boldsymbol{\kappa}_{\cdot,n}(\mathbf{x}) \mathbf{K}_{n,n}^{-1} \mathbf{f}_n$.
Define $\alpha_\text{aTS}(\mathbf{x})  = \frac{1}{N_\text{c}} \sum_{j=1}^{N_\text{c}} \widetilde{f}^j_\omega(\mathbf{x})$
as the sample-average posterior function, where $\widetilde{f}^j_\omega(\mathbf{x})$ are samples generated from the GP posterior
and $N_\text{c} \in \mathbb{N}_{>0}$. Since $\meanpost(\mathbf{x})$ is deterministic,
and the scaled prior sample $\frac{1}{\sqrt{N_\text{c}}} f^j_\omega(\mathbf{x})$ can be written as
$\frac{1}{\sqrt{N_\text{c}}} f^j_\omega(\mathbf{x}) \overset{\text{iid}}{\sim} \mathcal{GP}(0,\frac{1}{N_\text{c}} \kappa(\mathbf{x},\mathbf{x}'))$,
we have $\alpha_\text{aTS}(\mathbf{x})  = \mu(\mathbf{x}) + \frac{1}{\sqrt{N_\text{c}}} \left( \fprior(\mathbf{x}) + \xi_\omega(\mathbf{x})\right)$, where the first and second terms favor exploitation and exploration, respectively.
Thus, we can consider $N_\text{c}$ as an exploration--exploitation control parameter that, at large values, prioritizes exploitation
by concentrating the conditional distribution of the global minimum location, i.e., $p(\mathbf{x}^\star|\mathcal{D})$,
at the minimum location of $\meanpost(\mathbf{x})$, see \Cref{fig:averTSConcept} in \Cref{apd:AdditionalResults}.
With $\alpha_\text{aTS}(\mathbf{x})$, we can reproduce $\widetilde{f}_\omega(\mathbf{x})$ and the GP mean function $\meanpost(\mathbf{x})$
by setting $N_\text{c} = 1$ and  $N_\text{c} = \infty$, respectively.

\section{Related Works}

\paragraph{Sampling from Gaussian Process Posteriors.}
A prevalent method to sample GP posteriors with stationary covariance functions is via weight-space approximations based on Bayesian linear models of random Fourier features \citep{Rahimi2007}.
This method, unfortunately, is subject to the variance starvation problem \citep{Mutny2018,Wilson2020} which can be mitigated using more accurate feature representations (see e.g., \cite{Hensman2018,Solin2020}). An alternative is pathwise conditioning \citep{Wilson2020} that draws GP posterior samples by updating the corresponding prior samples. The decoupled representation of the pathwise conditioning can be further reformulated as two stochastic optimization problems for the posterior mean and an uncertainty reduction term, which are then efficiently solved using stochastic gradient descent \citep{LinJ2023}.

\paragraph{Optimization of Acquisition Functions.}
While their global optima guarantee the Bayes’ decision rule, BO acquisition functions are highly non-convex and difficult to optimize \citep{Wilson2018}.
Nevertheless, less attention has been given to the development of robust algorithms for optimizing these acquisition functions.
For this inner-loop optimization, gradient-based optimizers are often selected because of their fast convergence and robust performance \citep{Daulton2020}.
The implementation of such optimizers is facilitated by Monte Carlo (MC) acquisition functions whose derivatives are easy to evaluate \citep{Wilson2018}.
Gradient-based optimizers also allow multistart settings that use a set of starting points which can be, for example,
midpoints of data points \citep{Jones2001}, uniformly distributed samples over the input variable space \citep{Frazier2018,Ament2023},
or random points from a Latin hypercube design \citep{WangJ2020}.
However, multistart-based methods with random search may have difficulty determining the non-flat regions of acquisition functions, especially in high dimensions \citep{Rana2017}.
The log reformulation approach is a good solution to the numerical pathology of flat acquisition surface over large regions of the input variable space \citep{Ament2023}. While this approach works for acquisition functions prone to the flat surface issue such as the family of EI-based acquisition functions, its performance has yet to be evaluated for acquisition functions with many local minima like those based on posterior samples.

\paragraph{Posterior Sample-Based Acquisition Functions.}
As discussed in \Cref{introduction}, the family of posterior sample-based acquisition functions is determined from samples of the posterior.
GP-TS \citep{Chowdhury2017} is a notable member that extends the classical TS for finite-armed bandits to continuous settings of BO
(see algorithms in \Cref{apd:gp-ts}).
GP-TS prefers exploration by the mechanism that iteratively samples a function from the GP posterior of the objective function, optimizes this function, and selects the resulting solution as the next candidate for objective evaluation.
To further improve the exploitation of GP-TS, the sample mean of MC acquisition functions can be defined from multiple samples of the posterior \citep{Wilson2018,Balandat2020}. 
Different types of MC acquisition functions can also be used to inject beliefs about functions into the prior \citep{Hvarfner2024iclr}.

\section{Results}
\label{results}

We assess the performance of TS-roots in optimizing benchmark functions. We then compare the quality of solutions to the inner-loop optimization of GP-TS acquisition functions obtained from our proposed method, a gradient-based multistart optimizer with uniformly random starting points, and a genetic algorithm. 
We also show how TS-roots can improve the performance of MES.
Finally, we propose a new sample-average posterior function and show how it affects the performance of GP-TS. The experimental details for the presented results are in \Cref{apd:ExperimentDetails}.

\paragraph{Optimizing Benchmark Functions.}

We test the empirical performance of TS-roots on challenging minimization problems of five benchmark functions: the 2D Schwefel, 4D Rosenbrock, 10D Levy, 16D Ackley, and 16D Powell functions \citep{Surjanovic2013}.
The analytical expressions for these functions and their global minimum are given in \Cref{apd:BenchmarkFunctions}. 

In each optimization iteration, we record the best observed value of the error $\log(y_{\min} - f^\star)$
and the distance $\log\left(\left\| \mathbf{x}_{\min} - \mathbf{x}^\star\right\|\right)$,
where $y_{\min}$, $\mathbf{x}_{\min}$, $f^\star$, and $\mathbf{x}^\star$
are the best observation of the objective function in each iteration, the corresponding location of the observation,
the true minimum value of the objective function, and the true minimum location, respectively.
We compare the optimization results obtained from TS-roots and other BO methods, including GP-TS using decoupling sampling with random Fourier features (TS-DSRF), GP-TS with random Fourier features (TS-RF), expected improvement (EI) \citep{Jones1998}, lower confidence bound (LCB)---the version of GP-UCB \citep{Srinivas2010} for minimization problems, and LogEI \citep{Ament2023}.

\begin{figure}[t]
\centering
\includegraphics[width=\textwidth]{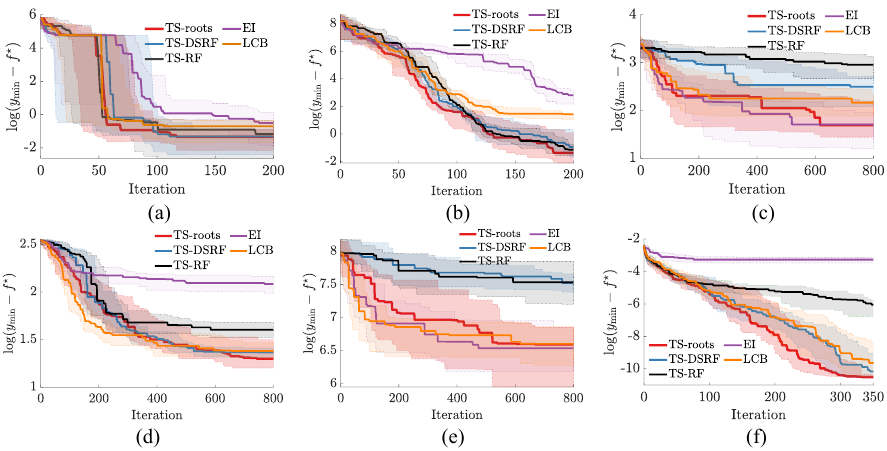}
\caption{Outer-loop optimization results for the (a) 2D Schwefel, (b) 4D Rosenbrock, (c) 10D Levy, (d) 16D Ackley, and (e) 16D Powell functions.
The plots are histories of medians and interquartile ranges of solution values from 20 runs of TS-roots, TS-DSRF
(i.e., TS using decoupled sampling with random Fourier features), TS-RF (i.e., TS using random Fourier features), EI, and LCB.} 
\label{fig:OptResultsValues}
\end{figure}

\Cref{fig:OptResultsValues} shows the medians and interquartile ranges of solution values obtained from 20 runs of each of the considered BO methods. The corresponding histories of solution locations are in \Cref{fig:OptResultsLocations} of \Cref{apd:AdditionalResults}.
With a fair comparison of outer-loop results (detailed in \Cref{apd:ExperimentDetails}),
TS-roots surprisingly shows the best performance on the 2D Schwefel, 16D Ackley, and 16D Powell functions,
and gives competitive results in the 4D Rosenbrock and 10D Levy problems.
Notably, TS-roots recommends better solutions than its counterparts, TS-DSRF and TS-RF, in high-dimensional problems
and offers competitive performance in low-dimensional problems.
Across all the examples, EI and LCB tend to perform better in the initial stages, while TS-roots shows fast improvement in later stages. This is because GP-TS favors exploration, which delays rewards.
The exploration phase, in general, takes longer for higher-dimensional problems.
TS-roots outperforms LogEI on the 2D Schwefel, 4D Rosenbrock, 10D Levy, and 16D Ackley functions (detailed in \Cref{fig:outerTSLogEI} of \Cref{apd:AdditionalResults}).

\paragraph{Optimizing Real-world Problem.}
We implement TS-roots to optimize an engineered ten-bar truss structure (see \Cref{apd:10bartruss}). Ten design variables of the truss are the cross-sectional areas of the truss members. The objective is to minimize a weighted sum of the scaled total cross-sectional area and the scaled vertical displacement at a node of interest.

Figures \ref{fig:OptResultsValues}(f) and \ref{fig:outerTSLogEI}(f) show the outer-loop optimization results for the truss obtained from 20 runs of each BO method, where $f^{\star}$ is a lower bound of the best objective function value we observed from all runs. TS-roots provides the best optimization result with rapid convergence.

\begin{figure}[t]
	\centering
	\includegraphics[width=\textwidth]{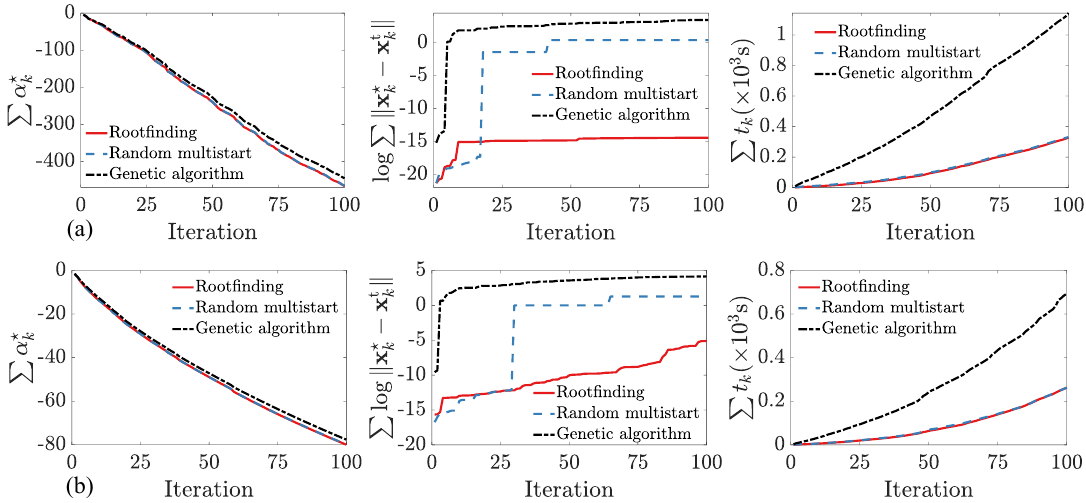}
	\caption{Inner-loop optimization results by rootfinding, a gradient-based multistart optimizer with random starting points (random multistart), and a genetic algorithm for (a) the 2D Schwefel and (b) 4D Rosenbrock functions. The plots are cumulative values of optimized GP-TS acquisition functions $\alpha_k^\star$, cumulative distances between new solution points ${\bf x}_k^\star$ and the true global minima ${\bf x}_k^\text{t}$ of the acquisition functions, and cumulative CPU times $t_k$ for optimizing the acquisition functions.} 
	\label{fig:innercompare1}
\end{figure}

\begin{figure}[t]
	\centering
	\includegraphics[width=\textwidth]{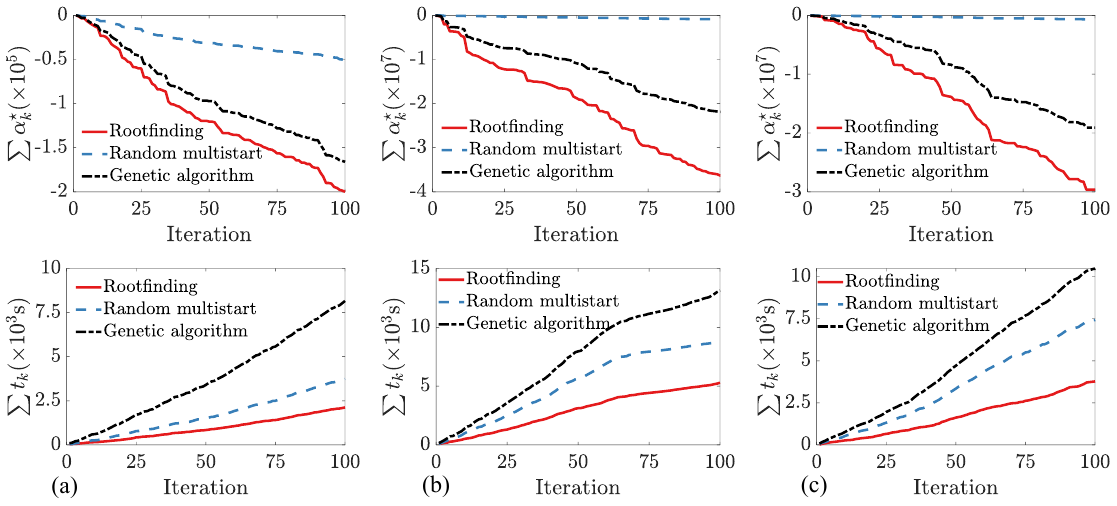}
	\caption{Inner-loop optimization results by rootfinding, a gradient-based multistart optimizer with random starting points (random multistart), and a genetic algorithm for (a) the 10D Levy, (b) 16D Ackley, and (c) 16D Powell functions. The plots are cumulative values of optimized GP-TS acquisition functions $\alpha_k^\star$ and cumulative CPU times $t_k$ for optimizing the acquisition functions.} 
	\label{fig:innercompare2}
\end{figure}

\paragraph{Optimizing GP-TS Acquisition Functions via Rootfinding.}

We assess the quality of solutions and computational cost for the inner-loop optimization of GP-TS acquisition functions by the proposed global optimization algorithm, referred to as rootfinding hereafter.
We do so by computing the optimized values $\alpha_k^\star$ of the GP-TS acquisition functions, the corresponding solution points $\mathbf{x}_k^\star$, and the CPU times $t_k$ required for optimizing the acquisition functions during the optimization process.
For low-dimensional problems of the 2D Schwefel and 4D Rosenbrock functions,
we also compute the exact global solution points $\mathbf{x}_k^\mathrm{t}$ of the GP-TS acquisition functions
by starting the gradient-based optimizer at a large number of initial points (set as $10^4$),
which is much larger than the maximum number of starting points set for TS-roots.
For comparison, we extend the same GP-TS acquisition functions to inner-loop optimization using a gradient-based multistart optimizer with random starting points (i.e., random multistart) and a genetic algorithm. 
In each outer-loop optimization iteration, the number of starting points for the random multistart and the population size of the genetic algorithm are equal to the number of starting points recommended for rootfinding.
The same termination conditions are used for the three algorithms.

\Cref{fig:innercompare1} shows the comparative performance of the inner-loop optimization for low-dimensional cases:
the 2D Schwefel and 4D Rosenbrock functions.
We see that the optimized acquisition function values and the optimization runtimes by rootfinding and the random multistart algorithm are almost identical, both of which are much better than those by the genetic algorithm. Rootfinding gives the best quality of the new solution points in both cases, while the genetic algorithm gives the worst. Notably in higher-dimensional settings of the 10D Levy, 16D Ackley, and 16D Powell functions shown in \Cref{fig:innercompare2}, rootfinding performs much better than the random multistart and genetic algorithm in terms of optimized acquisition values and optimization runtimes, which verifies the importance of the judicious selection of starting points for global optimization of the GP-TS acquisition functions and the efficiency of rootfinding in high dimensions.
The performance of random multistart becomes worse in higher dimensions.
\Cref{apd:AdditionalResults} provides additional results for gradient-based multistart optimization
using two other initialization schemes: uniform grid and Latin hypercube sampling.
Rootfinding outperforms both, especially in higher dimensions.

\paragraph{TS-roots to Information-Theoretic Acquisition Functions.}

We show how TS-roots can enhance the performance of MES \citep{WangZ2017},
which uses information about the maximum function value $f^\star$ for conducting BO.
One approach to computing MES generates a set of GP posterior samples using TS-RF
and subsequently optimizes the generated functions for samples of $f^\star$
using a gradient-based multistart optimizer with a large number of random starting points \citep{WangZ2017}.
We hypothesize that high-quality $f^\star$ samples can improve the performance of MES.
Thus, we assign both TS-roots and TS-RF as the inner workings of MES and then compare the resulting optimal solutions.
Note that the inner-loop optimization of MES, which strongly influences the optimization results, is not addressed by TS-roots.

Specifically, we minimize the 4D Rosenbrock, 6D Hartmann, and 10D Levy functions using four versions of MES, namely MES-R 10,  MES-R 50, MES-TS-roots 10, and MES-TS-roots 50.
Here, MES-R \citep{WangZ2017} and MES-TS-roots correspond to TS-RF and TS-roots, respectively, while 10 and 50 represent the number of random samples $f^\star$ generated for computing the MES acquisition function in each iteration.

\begin{figure}[t]
	\centering
	\includegraphics[width=\textwidth]{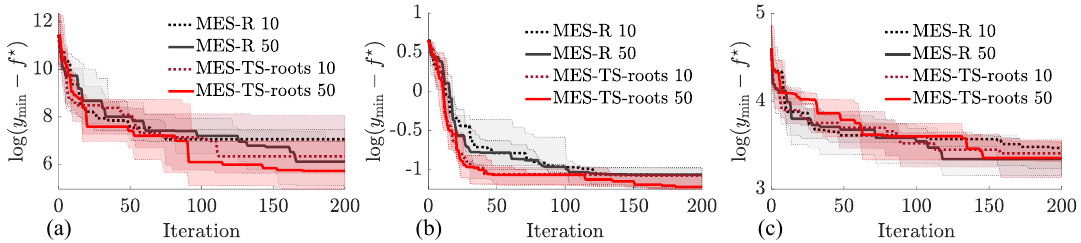}
	\caption{Performance of MES-R 10 and MES-R 50 for (a) the 4D Rosenbrock function, (b) the 6D Hartmann function, and (c) the 10D Levy function when TS-RF and TS-roots are used for generating random samples from $f^\star|\mathcal{D}$. The plots are histories of medians and interquartile ranges of solutions from ten runs of each method.} 
	\label{fig:MESResults}
\end{figure}

\Cref{fig:MESResults} shows the optimization histories for ten independent runs of each MES method.
On the 4D Rosenbrock and 6D Hartmann functions,
MES with TS-roots demonstrates superior optimization performance and faster convergence compared to MES with TS-RF,
especially when 50 samples of $f^\star$ are generated.
For the 10D Levy function, TS-roots outperforms TS-RF when using 10 samples of $f^\star$,
while their performance is comparable when 50 samples are used.

\paragraph{Performance of Sample-Average Posterior Functions.}

We investigate how $\alpha_\text{aTS}(\mathbf{x})$ influences the outer-loop optimization results. 
For this, we set $N_\text{c} \in \{1,10,50,100\}$ for TS-roots to optimize the 2D Schwefel, 4D Rosenbrock, and 6D Ackley functions.
We observe that increasing $N_\text{c}$ from $1$ to $10$ improves TS-roots performance on the 2D Schwefel, 4D Rosenbrock, and 6D Ackley functions (see \Cref{fig:averTSResults} in \Cref{apd:AdditionalResults}).
However, further increases in $N_\text{c}$ from 10 to 50 and 100 result in slight declines in solution quality as TS-roots transitions to exploitation.
These observations indicate that there is an optimal value of $N_\text{c}$ for each problem at which TS-roots achieves its best performance by balancing exploitation and exploration.
However, identifying the optimal value to maximize the performance of $\alpha_\text{aTS}(\mathbf{x})$
for a particular optimization problem remains an open issue.

\section{Conclusion and Future Work}

We presented TS-roots, a global optimization strategy for posterior sample paths.
It features an adaptive selection of starting points for gradient-based multistart optimizers,
combining exploration and exploitation.
This strategy breaks the curse of dimensionality by exploiting the separability of Gaussian process priors.
Compared with random multistart and a genetic algorithm, TS-roots consistently yields higher-quality solutions
in optimizing posterior sample paths, across a range of input dimensions.
It also improves the outer-loop optimization performance of GP-TS and information-theoretic acquisition functions such as MES
for Bayesian optimization. For future work, we aim to extend TS-roots to other spectral representations per Bochner’s theorem \citep{Mutny2018,Hensman2018,Solin2020}.
We also plan to study the ways and the probability of TS-roots failing to find the global optimum,
as well as the impact of subset sizes.

\subsubsection*{Acknowledgments}
We would like to thank Tian PAN for his help with the \texttt{maxk\_sum} problem,
and the organizers and the participants of the NeurIPS 2024 Workshop on Bayesian Decision-making and Uncertainty
for their invaluable discussions and feedback on an earlier version of the paper.
The authors are supported by the University of Houston through the SEED program no. 000189862.

\clearpage

\bibliographystyle{unsrtnat}
\bibliography{TS-roots} 

\begin{thebibliography}{53}
\providecommand{\natexlab}[1]{#1}
\providecommand{\url}[1]{\texttt{#1}}
\expandafter\ifx\csname urlstyle\endcsname\relax
  \providecommand{\doi}[1]{doi: #1}\else
  \providecommand{\doi}{doi: \begingroup \urlstyle{rm}\Url}\fi

\bibitem[Jones et~al.(1998)Jones, Schonlau, and Welch]{Jones1998}
Donald~R. Jones, Matthias Schonlau, and William~J. Welch.
\newblock Efficient global optimization of expensive black-box functions.
\newblock \emph{Journal of Global Optimization}, 13\penalty0 (4):\penalty0
  455--492, 1998.
\newblock \doi{10.1023/A:1008306431147}.
\newblock URL \url{https://doi.org/10.1023/A:1008306431147}.

\bibitem[Snoek et~al.(2012)Snoek, Larochelle, and Adams]{Snoek2012}
Jasper Snoek, Hugo Larochelle, and Ryan~P Adams.
\newblock Practical {B}ayesian optimization of machine learning algorithms.
\newblock In \emph{Advances in Neural Information Processing Systems},
  volume~25, pages 2951--2959, 2012.
\newblock URL
  \url{https://proceedings.neurips.cc/paper_files/paper/2012/file/05311655a15b75fab86956663e1819cd-Paper.pdf}.

\bibitem[Frazier(2018)]{Frazier2018}
Peter~I. Frazier.
\newblock Bayesian optimization.
\newblock In \emph{Recent Advances in Optimization and Modeling of Contemporary
  Problems}, {INFORMS TutORials in Operations Research}, chapter~11, pages
  255--278. October 2018.
\newblock \doi{10.1287/educ.2018.0188}.

\bibitem[Garnett(2023)]{Garnett2023}
Roman Garnett.
\newblock \emph{Bayesian Optimization}.
\newblock Cambridge University Press, Cambridge, UK, 2023.
\newblock \doi{10.1017/9781108348973}.

\bibitem[Srinivas et~al.(2010)Srinivas, Krause, Kakade, and
  Seeger]{Srinivas2010}
Niranjan Srinivas, Andreas Krause, Sham~M. Kakade, and Matthias Seeger.
\newblock {Gaussian process optimization in the bandit setting: No regret and
  experimental design}.
\newblock In \emph{Proceedings of the 27th International Conference on Machine
  Learning}, volume~13, pages 1015--1022, 2010.
\newblock URL \url{https://icml.cc/Conferences/2010/papers/422.pdf}.

\bibitem[Bull(2011)]{Bull2011}
Adam~D. Bull.
\newblock Convergence rates of efficient global optimization algorithms.
\newblock \emph{Journal of Machine Learning Research}, 12\penalty0
  (88):\penalty0 2879--2904, 2011.
\newblock URL \url{http://jmlr.org/papers/v12/bull11a.html}.

\bibitem[Russo and Van~Roy(2014)]{Russo2014}
Daniel Russo and Benjamin Van~Roy.
\newblock Learning to optimize via posterior sampling.
\newblock \emph{Mathematics of Operations Research}, 39\penalty0 (4):\penalty0
  1221--1243, April 2014.
\newblock \doi{10.1287/moor.2014.0650}.

\bibitem[Chowdhury and Gopalan(2017)]{Chowdhury2017}
Sayak~Ray Chowdhury and Aditya Gopalan.
\newblock On kernelized multi-armed bandits.
\newblock In \emph{Proceedings of the 34th International Conference on Machine
  Learning}, volume~70, pages 844--853, 06--11 Aug 2017.
\newblock URL \url{https://proceedings.mlr.press/v70/chowdhury17a.html}.

\bibitem[Wilson et~al.(2018)Wilson, Hutter, and Deisenroth]{Wilson2018}
James Wilson, Frank Hutter, and Marc Deisenroth.
\newblock Maximizing acquisition functions for {B}ayesian optimization.
\newblock In \emph{Advances in Neural Information Processing Systems},
  volume~31, pages 9884--9895, 2018.
\newblock URL
  \url{https://proceedings.neurips.cc/paper/2018/hash/498f2c21688f6451d9f5fd09d53edda7-Abstract.html}.

\bibitem[Rana et~al.(2017)Rana, Li, Gupta, Nguyen, and Venkatesh]{Rana2017}
Santu Rana, Cheng Li, Sunil Gupta, Vu~Nguyen, and Svetha Venkatesh.
\newblock High dimensional {B}ayesian optimization with elastic {G}aussian
  process.
\newblock In \emph{Proceedings of the 34th International Conference on Machine
  Learning}, volume~70, pages 2883--2891, 06--11 Aug 2017.
\newblock URL \url{https://proceedings.mlr.press/v70/rana17a.html}.

\bibitem[Hennig and Schuler(2012)]{Hennig2012}
Philipp Hennig and Christian~J. Schuler.
\newblock Entropy search for information-efficient global optimization.
\newblock \emph{Journal of Machine Learning Research}, 13\penalty0
  (6):\penalty0 1809--1837, 2012.
\newblock URL \url{https://www.jmlr.org/papers/v13/hennig12a.html}.

\bibitem[Hernández-Lobato et~al.(2014)Hernández-Lobato, Hoffman, and
  Ghahramani]{HernandezLobato2014}
José~Miguel Hernández-Lobato, Matthew~W. Hoffman, and Zoubin Ghahramani.
\newblock Predictive entropy search for efficient global optimization of
  black-box functions.
\newblock In \emph{Advances in Neural Information Processing Systems},
  volume~27, pages 918--926, 2014.
\newblock URL
  \url{https://proceedings.neurips.cc/paper_files/paper/2014/hash/069d3bb002acd8d7dd095917f9efe4cb-Abstract.html}.

\bibitem[Wang and Jegelka(2017)]{WangZ2017}
Zi~Wang and Stefanie Jegelka.
\newblock Max-value entropy search for efficient {B}ayesian optimization.
\newblock In \emph{Proceedings of the 34th International Conference on Machine
  Learning}, volume~70, pages 3627--3635, 2017.
\newblock URL \url{https://proceedings.mlr.press/v70/wang17e.html}.

\bibitem[Tu et~al.(2022)Tu, Gandy, Kantas, and Shafei]{Tu2022}
Ben Tu, Axel Gandy, Nikolas Kantas, and Behrang Shafei.
\newblock Joint entropy search for multi-objective bayesian optimization.
\newblock In \emph{Advances in Neural Information Processing Systems},
  volume~35, pages 9922--9938, 2022.
\newblock URL
  \url{https://proceedings.neurips.cc/paper_files/paper/2022/file/4086fe59dc3584708468fba0e459f6a7-Paper-Conference.pdf}.

\bibitem[Hvarfner et~al.(2022)Hvarfner, Hutter, and Nardi]{Hvarfner2022}
Carl Hvarfner, Frank Hutter, and Luigi Nardi.
\newblock Joint entropy search for maximally-informed {B}ayesian optimization.
\newblock In \emph{Advances in Neural Information Processing Systems},
  volume~35, pages 11494--11506, 2022.
\newblock URL
  \url{https://proceedings.neurips.cc/paper_files/paper/2022/hash/4b03821747e89ce803b2dac590f6a39b-Abstract-Conference.html}.

\bibitem[Russo and Van~Roy(2016)]{Russo2016}
Daniel~J. Russo and Benjamin Van~Roy.
\newblock An information-theoretic analysis of {T}hompson sampling.
\newblock \emph{The Journal of Machine Learning Research}, 17\penalty0
  (1):\penalty0 2442--2471, 2016.
\newblock URL \url{https://www.jmlr.org/papers/v17/14-087.html}.

\bibitem[Mutny and Krause(2018)]{Mutny2018}
Mojmir Mutny and Andreas Krause.
\newblock Efficient high dimensional {B}ayesian optimization with additivity
  and quadrature {F}ourier features.
\newblock In \emph{Advances in Neural Information Processing Systems},
  volume~31, pages 9005--9016, 2018.
\newblock URL
  \url{https://proceedings.neurips.cc/paper/2018/hash/4e5046fc8d6a97d18a5f54beaed54dea-Abstract.html}.

\bibitem[MacKay(2003)]{MacKay2003}
David J.~C. MacKay.
\newblock \emph{Information theory, inference, and learning algorithms}.
\newblock Cambridge University Press, Cambridge, UK, 2003.
\newblock URL \url{https://www.cambridge.org/9780521642989}.

\bibitem[Shah and Ghahramani(2015)]{Shah2015}
Amar Shah and Zoubin Ghahramani.
\newblock Parallel predictive entropy search for batch global optimization of
  expensive objective functions.
\newblock In \emph{Advances in Neural Information Processing Systems},
  volume~28, 2015.
\newblock URL
  \url{https://proceedings.neurips.cc/paper_files/paper/2015/file/57c0531e13f40b91b3b0f1a30b529a1d-Paper.pdf}.

\bibitem[Hern{\'a}ndez-Lobato et~al.(2017)Hern{\'a}ndez-Lobato, Requeima,
  Pyzer-Knapp, and Aspuru-Guzik]{HernandezLobato2017}
Jos{\'e}~Miguel Hern{\'a}ndez-Lobato, James Requeima, Edward~O. Pyzer-Knapp,
  and Al{\'a}n Aspuru-Guzik.
\newblock Parallel and distributed {T}hompson sampling for large-scale
  accelerated exploration of chemical space.
\newblock In \emph{Proceedings of the 34th International Conference on Machine
  Learning}, volume~70, pages 1470--1479, 06--11 Aug 2017.
\newblock URL \url{https://proceedings.mlr.press/v70/hernandez-lobato17a.html}.

\bibitem[Kandasamy et~al.(2018)Kandasamy, Krishnamurthy, Schneider, and
  Poczos]{Kandasamy2018}
Kirthevasan Kandasamy, Akshay Krishnamurthy, Jeff Schneider, and Barnabas
  Poczos.
\newblock Parallelised {B}ayesian optimisation via {T}hompson sampling.
\newblock In \emph{Proceedings of the Twenty-First International Conference on
  Artificial Intelligence and Statistics}, volume~84, pages 133--142, 09--11
  Apr 2018.
\newblock URL \url{https://proceedings.mlr.press/v84/kandasamy18a.html}.

\bibitem[Kushner(1964)]{Kushner1964}
Harold~J. Kushner.
\newblock A new method of locating the maximum point of an arbitrary multipeak
  curve in the presence of noise.
\newblock \emph{Journal Basic Engineering}, 86\penalty0 (1):\penalty0 97--106,
  1964.
\newblock \doi{10.1115/1.3653121}.

\bibitem[Jones et~al.(1993)Jones, Perttunen, and Stuckman]{Jones1993}
D.~R. Jones, C.~D. Perttunen, and B.~E. Stuckman.
\newblock Lipschitzian optimization without the {L}ipschitz constant.
\newblock \emph{Journal of Optimization Theory and Applications}, 79\penalty0
  (1):\penalty0 157--181, 1993.
\newblock \doi{10.1007/BF00941892}.
\newblock URL \url{https://doi.org/10.1007/BF00941892}.

\bibitem[Hansen et~al.(2003)Hansen, Müller, and Koumoutsakos]{Hansen2003}
Nikolaus Hansen, Sibylle~D. Müller, and Petros Koumoutsakos.
\newblock Reducing the time complexity of the derandomized evolution strategy
  with covariance matrix adaptation {(CMA-ES)}.
\newblock \emph{Evolutionary Computation}, 11\penalty0 (1):\penalty0 1--18,
  2003.
\newblock \doi{10.1162/106365603321828970}.

\bibitem[Mitchell(1998)]{Mitchell1998}
Melanie Mitchell.
\newblock \emph{An introduction to genetic algorithms}.
\newblock The MIT Press, Massachusetts, USA, 1998.

\bibitem[Kim and Choi(2021)]{Kim2021}
Jungtaek Kim and Seungjin Choi.
\newblock On local optimizers of acquisition functions in {Bayesian}
  optimization.
\newblock In \emph{Machine Learning and Knowledge Discovery in Databases},
  pages 675--690, 2021.
\newblock \doi{10.1007/978-3-030-67661-2_40}.

\bibitem[Bergstra and Bengio(2012)]{Bergstra2012}
James Bergstra and Yoshua Bengio.
\newblock Random search for hyper-parameter optimization.
\newblock \emph{Journal of Machine Learning Research}, 13\penalty0
  (10):\penalty0 281--305, 2012.
\newblock URL \url{http://jmlr.org/papers/v13/bergstra12a.html}.

\bibitem[Balandat et~al.(2020)Balandat, Karrer, Jiang, Daulton, Letham, Wilson,
  and Bakshy]{Balandat2020}
Maximilian Balandat, Brian Karrer, Daniel Jiang, Samuel Daulton, Ben Letham,
  Andrew~G Wilson, and Eytan Bakshy.
\newblock {BoTorch: A framework for efficient Monte-Carlo Bayesian
  optimization}.
\newblock In \emph{Advances in Neural Information Processing Systems},
  volume~33, pages 21524--21538, 2020.
\newblock URL
  \url{https://proceedings.neurips.cc/paper_files/paper/2020/file/f5b1b89d98b7286673128a5fb112cb9a-Paper.pdf}.

\bibitem[Feo and Resende(1995)]{Feo1995}
Thomas~A. Feo and Mauricio G.~C. Resende.
\newblock Greedy randomized adaptive search procedures.
\newblock \emph{Journal of Global Optimization}, 6\penalty0 (2):\penalty0
  109--133, 1995.
\newblock \doi{10.1007/BF01096763}.
\newblock URL \url{https://doi.org/10.1007/BF01096763}.

\bibitem[Chapelle and Li(2011)]{Chapelle2011}
Olivier Chapelle and Lihong Li.
\newblock An empirical evaluation of {T}hompson sampling.
\newblock In \emph{Advances in Neural Information Processing Systems},
  volume~24, pages 2249--2257, 2011.
\newblock URL
  \url{https://papers.nips.cc/paper_files/paper/2011/hash/e53a0a2978c28872a4505bdb51db06dc-Abstract.html}.

\bibitem[Wilson et~al.(2020)Wilson, Borovitskiy, Terenin, Mostowsky, and
  Deisenroth]{Wilson2020}
James Wilson, Viacheslav Borovitskiy, Alexander Terenin, Peter Mostowsky, and
  Marc Deisenroth.
\newblock Efficiently sampling functions from {G}aussian process posteriors.
\newblock In \emph{Proceedings of the 37th International Conference on Machine
  Learning}, volume 119, pages 10292--10302, 13--18 Jul 2020.
\newblock URL \url{https://proceedings.mlr.press/v119/wilson20a.html}.

\bibitem[Wilson et~al.(2021)Wilson, Borovitskiy, Terenin, Mostowsky, and
  Deisenroth]{Wilson2021}
James~T. Wilson, Viacheslav Borovitskiy, Alexander Terenin, Peter Mostowsky,
  and Marc~Peter Deisenroth.
\newblock Pathwise conditioning of gaussian processes.
\newblock \emph{Journal of Machine Learning Research}, 22\penalty0
  (105):\penalty0 1--47, 2021.
\newblock URL \url{http://jmlr.org/papers/v22/20-1260.html}.

\bibitem[{Hoffman} and {Ribak}(1991)]{Hoffman1991}
Yehuda {Hoffman} and Erez {Ribak}.
\newblock {Constrained Realizations of Gaussian Fields: A Simple Algorithm}.
\newblock \emph{Astrophysical Journal Letters}, 380:\penalty0 L5--L8, October
  1991.
\newblock \doi{10.1086/186160}.
\newblock URL \url{https://ui.adsabs.harvard.edu/abs/1991ApJ...380L...5H}.

\bibitem[Lin et~al.(2023)Lin, Antor\'{a}n, Padhy, Janz, Hern\'{a}ndez-Lobato,
  and Terenin]{LinJ2023}
Jihao~Andreas Lin, Javier Antor\'{a}n, Shreyas Padhy, David Janz,
  Jos\'{e}~Miguel Hern\'{a}ndez-Lobato, and Alexander Terenin.
\newblock Sampling from {G}aussian process posteriors using stochastic gradient
  descent.
\newblock In \emph{Advances in Neural Information Processing Systems},
  volume~36, pages 36886--36912, 2023.
\newblock URL
  \url{https://proceedings.neurips.cc/paper_files/paper/2023/file/7482e8ce4139df1a2d8195a0746fa713-Paper-Conference.pdf}.

\bibitem[Duvenaud et~al.(2013)Duvenaud, Lloyd, Grosse, Tenenbaum, and
  Zoubin]{Duvenaud2013}
David Duvenaud, James Lloyd, Roger Grosse, Joshua Tenenbaum, and Ghahramani
  Zoubin.
\newblock Structure discovery in nonparametric regression through compositional
  kernel search.
\newblock In \emph{Proceedings of the 30th International Conference on Machine
  Learning}, volume~28, pages 1166--1174, 2013.
\newblock URL \url{http://proceedings.mlr.press/v28/duvenaud13.html}.

\bibitem[Kanagawa et~al.(2018)Kanagawa, Hennig, Sejdinovic, and
  Sriperumbudur]{Kanagawa2018}
Motonobu Kanagawa, Philipp Hennig, Dino Sejdinovic, and Bharath~K
  Sriperumbudur.
\newblock Gaussian processes and kernel methods: A review on connections and
  equivalences, 2018.
\newblock URL \url{https://arxiv.org/abs/1807.02582}.

\bibitem[Trefethen(2019)]{Trefethen2019}
Lloyd~N. Trefethen.
\newblock \emph{Approximation Theory and Approximation Practice, Extended
  Edition}.
\newblock Society for Industrial and Applied Mathematics, Philadelphia, PA,
  2019.
\newblock ISBN 978-1-61197-593-2.
\newblock \doi{10.1137/1.9781611975949}.
\newblock URL \url{https://people.maths.ox.ac.uk/trefethen/ATAP/}.

\bibitem[Rahimi and Recht(2007)]{Rahimi2007}
Ali Rahimi and Benjamin Recht.
\newblock Random features for large-scale kernel machines.
\newblock In \emph{Advances in Neural Information Processing Systems},
  volume~20, pages 1177--1184, 2007.
\newblock URL
  \url{https://proceedings.neurips.cc/paper_files/paper/2007/file/013a006f03dbc5392effeb8f18fda755-Paper.pdf}.

\bibitem[Hensman et~al.(2018)Hensman, Durrande, and Solin]{Hensman2018}
James Hensman, Nicolas Durrande, and Arno Solin.
\newblock Variational {F}ourier features for {G}aussian processes.
\newblock \emph{Journal of Machine Learning Research}, 18\penalty0
  (151):\penalty0 1--52, 2018.
\newblock URL \url{http://jmlr.org/papers/v18/16-579.html}.

\bibitem[Solin and Särkkä(2020)]{Solin2020}
Arno Solin and Simo Särkkä.
\newblock Hilbert space methods for reduced-rank {G}aussian process regression.
\newblock \emph{Statistics and Computing}, 30\penalty0 (2):\penalty0 419--446,
  2020.
\newblock \doi{10.1007/s11222-019-09886-w}.

\bibitem[Daulton et~al.(2020)Daulton, Balandat, and Bakshy]{Daulton2020}
Samuel Daulton, Maximilian Balandat, and Eytan Bakshy.
\newblock Differentiable expected hypervolume improvement for parallel
  multi-objective bayesian optimization.
\newblock In \emph{Advances in Neural Information Processing Systems},
  volume~33, pages 9851--9864, 2020.
\newblock URL
  \url{https://proceedings.neurips.cc/paper_files/paper/2020/file/6fec24eac8f18ed793f5eaad3dd7977c-Paper.pdf}.

\bibitem[Jones(2001)]{Jones2001}
Donald~R. Jones.
\newblock A taxonomy of global optimization methods based on response surfaces.
\newblock \emph{Journal of Global Optimization}, 21\penalty0 (4):\penalty0
  345--383, 2001.
\newblock \doi{10.1023/A:1012771025575}.

\bibitem[Ament et~al.(2023)Ament, Daulton, Eriksson, Balandat, and
  Bakshy]{Ament2023}
Sebastian Ament, Samuel Daulton, David Eriksson, Maximilian Balandat, and Eytan
  Bakshy.
\newblock Unexpected improvements to expected improvement for {B}ayesian
  optimization.
\newblock In \emph{Advances in Neural Information Processing Systems},
  volume~36, pages 20577--20612, 2023.
\newblock URL
  \url{https://proceedings.neurips.cc/paper_files/paper/2023/file/419f72cbd568ad62183f8132a3605a2a-Paper-Conference.pdf}.

\bibitem[Wang et~al.(2020)Wang, Clark, Liu, and Frazier]{WangJ2020}
Jialei Wang, Scott~C. Clark, Eric Liu, and Peter~I. Frazier.
\newblock Parallel {B}ayesian global optimization of expensive functions.
\newblock \emph{Operations Research}, 68\penalty0 (6):\penalty0 1850--1865,
  2020.
\newblock \doi{10.1287/opre.2019.1966}.
\newblock URL \url{https://doi.org/10.1287/opre.2019.1966}.

\bibitem[Hvarfner et~al.(2024)Hvarfner, Hutter, and Nardi]{Hvarfner2024iclr}
Carl Hvarfner, Frank Hutter, and Luigi Nardi.
\newblock A general framework for user-guided bayesian optimization.
\newblock In \emph{The Twelfth International Conference on Learning
  Representations}, pages 9851--9864, 2024.
\newblock URL \url{https://iclr.cc/virtual/2024/poster/18774}.

\bibitem[Surjanovic and Bingham(2013)]{Surjanovic2013}
S~Surjanovic and D~Bingham.
\newblock Virtual library of simulation experiments: Test functions and
  datasets, 2013.
\newblock URL \url{http://www.sfu.ca/~ssurjano/optimization.html}.

\bibitem[Rasmussen and Williams(2006)]{Rasmussen2006}
Carl~Edward Rasmussen and Christopher K~I Williams.
\newblock \emph{Gaussian processes for machine learning}.
\newblock The MIT Press, Massachusetts, USA, 2006.
\newblock ISBN 9780521872508.
\newblock \doi{10.7551/mitpress/3206.001.0001}.
\newblock URL \url{https://doi.org/10.7551/mitpress/3206.001.0001}.

\bibitem[Zhu et~al.(1998)Zhu, Williams, Rohwer, and Morciniec]{ZhuHY1998}
Huaiyu Zhu, C.~K.~I Williams, R~Rohwer, and M~Morciniec.
\newblock Gaussian regression and optimal finite dimensional linear models.
\newblock In \emph{Neural Networks and Machine Learning}, 1998.
\newblock URL \url{https://publications.aston.ac.uk/id/eprint/38366/}.

\bibitem[Gradshteyn and Ryzhik(2014)]{Gradshteyn2014}
I.S. Gradshteyn and I.M. Ryzhik.
\newblock \emph{Table of Integrals, Series, and Products}.
\newblock Academic Press, Boston, 8th edition, 2014.
\newblock ISBN 978-0-12-384933-5.
\newblock \doi{10.1016/C2010-0-64839-5}.

\bibitem[Battles and Trefethen(2004)]{Battles2004}
Zachary Battles and Lloyd~N. Trefethen.
\newblock An extension of {MATLAB} to continuous functions and operators.
\newblock \emph{SIAM Journal on Scientific Computing}, 25\penalty0
  (5):\penalty0 1743--1770, 2004.
\newblock \doi{10.1137/S1064827503430126}.

\bibitem[Richardson(2016)]{Richardson2016}
Mark Richardson.
\newblock chebpy, a {P}ython implementation of chebfun, 2016.
\newblock URL \url{https://github.com/chebpy/chebpy}.

\bibitem[Picheny et~al.(2013)Picheny, Wagner, and Ginsbourger]{Picheny2013}
Victor Picheny, Tobias Wagner, and David Ginsbourger.
\newblock A benchmark of kriging-based infill criteria for noisy optimization.
\newblock \emph{Structural and multidisciplinary optimization}, 48:\penalty0
  607--626, 2013.
\newblock \doi{10.1007/s00158-013-0919-4}.
\newblock URL \url{https://doi.org/10.1007/s00158-013-0919-4}.

\bibitem[Owen(1992)]{Owen1992}
Art~B. Owen.
\newblock {A Central Limit Theorem for Latin Hypercube Sampling}.
\newblock \emph{Journal of the Royal Statistical Society: Series B
  (Methodological)}, 54\penalty0 (2):\penalty0 541--551, 12 1992.
\newblock \doi{10.1111/j.2517-6161.1992.tb01895.x}.
\newblock URL \url{https://doi.org/10.1111/j.2517-6161.1992.tb01895.x}.

\end{thebibliography}

\clearpage
\appendix

\section{Characterizing the Local Minima of a Separable Function}
\label{apd:proofs}

\subsection{Proof of \Cref{prop:representation_extrema}: A Representation of the Set of Local Minima}

\Cref{prop:representation_extrema} broadly applies to separable functions on a hypercube.
Consider a separable function $f(\mathbf{x}) = \prod_{i=1}^d f_i(x_i)$ with domain $\mathcal{X} = \prod_{i=1}^d [\ximin, \ximax]$,
where $f_i \in C^1([\ximin, \ximax]; \mathbb{R})$.
To simplify the discussion, we further assume that $f_i$ is twice differentiable at its interior critical points $\xicritical$.
The gradient of $f$ can be written as:
\begin{equation}\label{eq:gradient}
\nabla f(\mathbf{x}) = \Big( f'_i(x_i) \cdot \prod_{j \neq i} f_j(x_j) \Big)_{i=1}^d
= \bigg( f(\mathbf{x}) \cdot \frac{f'_i(x_i)}{f_i(x_i)} \bigg)_{i=1}^d
= f(\mathbf{x}) \cdot \mathbf{v}(\mathbf{x}),
\end{equation}
where $\mathbf{v}(\mathbf{x}) = \big(f'_i / f_i\big)_{i=1}^d = \big(\frac{\diff}{\diff x_i} \log f_i\big)_{i=1}^d$.
The Hessian of $f$ can be written as:
\begin{equation}\label{eq:Hessian}
\nabla^2 f(\mathbf{x}) = \diag\Big\{ f''_i(x_i) \prod_{j \neq i} f_j(x_j) \Big\}_{i= 1}^d
+ \Big[ f'_i(x_i) f'_j(x_j) \prod_{k \neq i,j} f_k(x_k) \Big]_{i \in d}^{j \neq i}
= f(\mathbf{x}) \diag(\mathbf{s} + \mathbf{v} \mathbf{v}^\intercal),
\end{equation}
where $\mathbf{s}(\mathbf{x}) = \big(f''_i / f_i - (f'_i / f_i)^2 \big)_{i=1}^d = \big(\frac{\diff^2}{\diff x_i^2} \log f_i\big)_{i=1}^d$.

An interior point $\mathbf{x} \in \interior \mathcal{X} := \prod_{i=1}^d (\ximin, \ximax)$ is a strong local minimum of $f$
if and only if $\nabla f(\mathbf{x}) = 0$ and $\nabla^2 f(\mathbf{x}) > 0$.
From \cref{eq:gradient}, the first condition is satisfied in any of the following three cases:
\begin{enumerate*}[label=(\arabic*)]
\item $f_i(x_i) \ne 0$ and $f'_i(x_i) = 0$ for all $i \in \dset$;
\item $f_i(x_i) = 0$ for exactly one $i \in \dset$ and $f'_i(x_i) = 0$; or
\item $f_i(x_i) = 0$ for all $i \in I \subseteq \dset$ where $|I| \ge 2$.
\end{enumerate*}

In case (1), the Hessian \cref{eq:Hessian} reduces to
$\nabla^2 f(\mathbf{x}) = f(\mathbf{x}) \cdot \diag\{f''_i(x_i) / f_i(x_i) \}_{i= 1}^d$,
which is positive definite if and only if one of the following holds:
\begin{enumerate*}[label=(\roman*)]
\item $f(\mathbf{x}) > 0$ and $f_i(x_i) f''_i(x_i) > 0$, for all $i \in \dset$; or
\item $f(\mathbf{x}) < 0$ and $f_i(x_i) f''_i(x_i) < 0$, for all $i \in \dset$.
\end{enumerate*}

In case (2), the Hessian reduces to an all-zero matrix except for the $i$th diagonal entry:
$[\nabla^2 f(\mathbf{x})]_{i,i} = f''_i(x_i) \prod_{j \neq i} f_j(x_j)$.
Even if this entry is positive, the Hessian is still positive semi-definite,
which means that there is a continuum of weak local minima: $\{x_i\} \times \prod_{j \ne i} [\xjmin, \xjmax]$.
Besides, this case requires $f_i$ and $f'_i$ to have an identical root, which an event with probability zero.

In case (3), let $g_i(r_i) := f_i(x_i + r_i)$ be a shifted version of $f_i$, $i \in \dset$.
Taylor expansion at $\mathbf{r} = 0$ gives $g_i(r_i) = 0 + g'_i(0) \, r_i + o(r_i)$ for all $i \in I$
and $g_j(r_j) = g_j(0) + O(r_j)$ for all $j \notin I$.
We have $g(\mathbf{r}) := \prod_{i=1}^d g_i(r_i) = c \prod_{i \in I} r_i + o(\prod_{i \in I} r_i) \cdot O(\prod_{j \notin I} r_j)$,
where $c = \prod_{i \in I} g'_i(0) \cdot \prod_{j \notin I} g_j(0) \ne 0$.
This means that there is a continuum of saddle points: $\{x_i\}_{i \in I} \times \prod_{j \notin I} [\xjmin, \xjmax]$.

For a boundary point $\mathbf{x} \in \partial \mathcal{X} := \mathcal{X} \setminus \interior \mathcal{X}$,
we partition the index set $\dset$ into $L, R$, and $I$ such that
$x_i = \ximin$ for all $i \in L$, $x_i = \ximax$ for all $i \in R$, and $x_i \in (\ximin, \ximax)$ for all $i \in I$.
Define $\nabla_J := (\partial_j)_{j \in J}$ for any subset $J$ of the indices.
Then $\mathbf{x}$ is a strong local minimum of $f$ if and only if the following conditions hold:
\begin{enumerate*}[label=(\alph*)]
\item\label{i:b1} $\mathbf{x}$ is a strong local minimum in $\{x_j\}_{j \notin I} \times \prod_{j \in I} [\xjmin, \xjmax]$;
\item\label{i:b2} $\nabla_L f(\mathbf{x}) > 0$; and
\item\label{i:b3} $\nabla_R f(\mathbf{x}) < 0$.
\end{enumerate*}

Condition \ref{i:b1} holds if any only if $\nabla_I f(\mathbf{x}) = 0$ and $\nabla_I^2 f(\mathbf{x}) > 0$.
Based on the previous discussion on interior local minima, it is equivalent to:
\begin{enumerate*}[label=(\roman*)]
\item $f(\mathbf{x}) > 0$ and $f_i(x_i) f''_i(x_i) > 0$, for all $i \in I$; or
\item $f(\mathbf{x}) < 0$ and $f_i(x_i) f''_i(x_i) < 0$, for all $i \in I$.
\end{enumerate*}

From \cref{eq:gradient}, condition \ref{i:b2} is equivalent to:
\begin{enumerate*}[label=(\roman*)]
\item $f(\mathbf{x}) > 0$ and $f_i(x_i) f'_i(x_i) > 0$, for all $i \in L$; or
\item $f(\mathbf{x}) < 0$ and $f_i(x_i) f'_i(x_i) < 0$, for all $i \in L$.
\end{enumerate*}

Similarly, condition \ref{i:b3} is equivalent to:
\begin{enumerate*}[label=(\roman*)]
\item $f(\mathbf{x}) > 0$ and $-f_i(x_i) f'_i(x_i) > 0$, for all $i \in R$; or
\item $f(\mathbf{x}) < 0$ and $-f_i(x_i) f'_i(x_i) < 0$, for all $i \in R$.
\end{enumerate*}

Summarizing the above discussions, we see that there is a unified way to identify the set $\Xlocmin$
of all strong local minima of $f$, which is stated in \Cref{prop:representation_extrema}.
The discussion for the set $\Xlocmax$ of local maxima is the exactly the same, except that the signs are flipped.
This also means that $\Xlocmax$ and $\Xlocmin$ form a partition of the union $\Xmono \sqcup \Xmixed$
of the two tensor grids.

If $f_i$ is not twice differentiable at some interior critical point $x_i$,
we may replace $f''_i(x_i) > 0$ with the statement that $x_i$ is a strong local minimum of $f_i$,
and replace $f''_i(x_i) < 0$ with the statement that $x_i$ is a strong local maximum of $f_i$.
The rest of the discussion still follows.
In practice, the differentiability of the prior sample is not an issue,
because it is almost always approximated by a finite sum of analytic functions, which is again analytic.

\subsection{Number of Local Minima of a Separable Function}

In \cref{prop:representation_extrema}, each set of candidate coordinates $\xicandidate$ is partitioned into
mixed type and mono type:
\begin{equation*}
\xicandidatemixed = \{\xi_{i,j} \in \xicandidate : f_i(\xi_{i,j}) h_i(\xi_{i,j}) < 0\}, \quad
\xicandidatemono = \{\xi_{i,j} \in \xicandidate : f_i(\xi_{i,j}) h_i(\xi_{i,j}) > 0\}.
\end{equation*}
Another partition of $\xicandidate$ is by the sign of the corresponding component function value:
\begin{equation*}
\xicandidateneg = \{\xi_{i,j} \in \xicandidate : f_i(\xi_{i,j}) < 0\}, \quad
\xicandidatepos = \{\xi_{i,j} \in \xicandidate : f_i(\xi_{i,j}) > 0\}.
\end{equation*}
These two partitions create a finer partition of $\xicandidate$ into four subsets:
\begin{equation*}
\xicandidatenegmixed = \xicandidateneg \cap \xicandidatemixed, \quad
\xicandidatenegmono = \xicandidateneg \cap \xicandidatemono, \quad
\xicandidateposmixed = \xicandidatepos \cap \xicandidatemixed, \quad
\xicandidateposmono = \xicandidatepos \cap \xicandidatemono.
\end{equation*}
Denote the sizes of mixed and mono type candidate coordinates as
$n_i^{(1)} = |\xicandidatemixed|$ and $n_i^{(0)} = |\xicandidatemono|$,
then the sizes of the two tensor grids $\Xmixed$ and $\Xmono$ can be written as:
\begin{equation*}
N^{(1)} := |\Xmixed| = \prod_{i=1}^d n_i^{(1)}, \quad
N^{(0)} := |\Xmono| = \prod_{i=1}^d n_i^{(0)}.
\end{equation*}
Define signed sums as the sums of signs of function values on the two tensor grids:
\begin{equation*}
S^{(1)} := \sum_{\boldsymbol{\xi} \in \Xmixed} \sign(f(\boldsymbol{\xi})), \quad
S^{(0)} := \sum_{\boldsymbol{\xi} \in \Xmono} \sign(f(\boldsymbol{\xi})).
\end{equation*}
We now derive efficient formulas to calculate these signed sums, using $S^{(1)}$ as an example.
Denote each coordinate in $\xicandidatemixed$ as $\xi_{i,j}^{(1)}$.
Denote each point in $\Xmixed$ as $\boldsymbol{\xi}_J^{(1)} = (\xi_{i,J_i}^{(1)})_{i=1}^d$,
where multi-index $J = (J_i)_{i=1}^d \in \Pi^{(1)} := \prod_{i=1}^d \{1, \cdots, n_i^{(1)}\}$.
The signed sum $S^{(1)}$ can be written as:
\begin{align*}
S^{(1)}
&= \sum_{J \in \Pi^{(1)}} \sign(f(\boldsymbol{\xi}_J^{(1)}))
= \sum_{J \in \Pi^{(1)}} \sign\bigg(\prod_{i=1}^d f_i(\xi_{i,J_i}^{(1)})\bigg) \\
&= \sum_{J \in \Pi^{(1)}} \prod_{i=1}^d \sign(f_i(\xi_{i,J_i}^{(1)}))
= \prod_{i=1}^d \sum_{j = 1}^{n_i^{(1)}}  \sign(f_i(\xi_{i,j}^{(1)})) \\
&= \prod_{i=1}^d \left[ \sum_{j = 1}^{n_i^{(1)}}  1(f_i(\xi_{i,j}^{(1)}) > 0) -
    \sum_{j = 1}^{n_i^{(1)}}  1(f_i(\xi_{i,j}^{(1)}) < 0) \right]
= \prod_{i=1}^d \left[ |\xicandidateposmixed| - |\xicandidatenegmixed| \right].
\end{align*}
A formula for $S^{(0)}$ can be derived analogously.
Denote set sizes:
\begin{equation*}
n_i^{-(1)} = |\xicandidatenegmixed|, \quad
n_i^{-(0)} = |\xicandidatenegmono|, \quad
n_i^{+(1)} = |\xicandidateposmixed|, \quad
n_i^{+(0)} = |\xicandidateposmono|,
\end{equation*}
then the signed sums can be calculated as:
\begin{equation*}
S^{(1)} = \prod_{i=1}^d (n_i^{+(1)} - n_i^{-(1)}), \quad
S^{(0)} = \prod_{i=1}^d (n_i^{+(0)} - n_i^{-(0)}).
\end{equation*}
The sizes of negative and positive strong local minima of a separable function can be written as:
\begin{align}
\breve{N}^{-} := |\Xneglocmin| = \sum_{\boldsymbol{\xi} \in \Xmixed} 1(f(\boldsymbol{\xi}) < 0)
= \frac{1}{2} (N^{(1)} - S^{(1)}), \\
\breve{N}^{+} := |\Xposlocmin| = \sum_{\boldsymbol{\xi} \in \Xmono} 1(f(\boldsymbol{\xi}) > 0)
= \frac{1}{2} (N^{(0)} + S^{(0)}). \nonumber
\end{align}
Therefore, the size of the strong local minima of a separable function can be written as:
\begin{equation}
\breve{N} := |\Xlocmin| = |\Xneglocmin| + |\Xposlocmin| = \frac{1}{2} (N^{(1)} + N^{(0)} - S^{(1)} + S^{(0)}).
\end{equation}

\section{Ordering the Local Minima of a Separable Function}
\label{apd:ordering}

\subsection{Filtering a Tensor Grid for High Absolute Values of a Separable Function}

The step one in \Cref{sec:ordering_minima} is equivalent to the following problem:
given coordinates $Z_i = \{\zeta_{i,1}, \cdots, \zeta_{i,t_i}\}$
and components values $F_i = \{f_{i,1}, \cdots, f_{i,t_i}\}$, $i \in \dset$,
of a separable function $f(\mathbf{x}) = \prod_{i=1}^d f_i(x_i)$,
find points $\boldsymbol{\zeta}$ such that $|f(\boldsymbol{\zeta})|$ are the $k$ largest
in the tensor grid $Z = \prod_{i=1}^d Z_i$.

Because $\log |f(\mathbf{x})| = \log |\prod_{i=1}^d f_i(x_i)| = \sum_{i=1}^d \log |f_i(x_i)|$,
we can solve this problem as follows:
define two-dimensional arrays $F = [F_1, \cdots, F_d]$ and $A = \log |F|$,
solve $S = \texttt{maxk\_sum}(A, k)$,
and return $\{\boldsymbol{\zeta} = (\zeta_{1,I_1}, \cdots, \zeta_{d,I_d}) : I \in S \}$.
Here the \texttt{maxk\_sum} algorithms finds the combinations from $A$ that gives the $k$ largest sums,
which is described next.

\subsection{Top Combinations with the Largest Sums}

Consider this problem: given a two-dimensional array $A = [\mathbf{a}_1, \cdots, \mathbf{a}_d]$,
$\mathbf{a}_i = [a_{i, 1}, \cdots, a_{i, t_i}]$, with $a_{i, 1} \ge \cdots \ge a_{i, t_i}$, $i \in \dset$,
find $k$ multi-indices of the form $I = [I_1, \cdots, I_d]$ such that the sums $s_I := \sum_{i=1}^d a_{i, I_i}$
are the $k$ largest among all combinations $I \in \prod_{i=1}^d \{1, \cdots, t_i\}$.

An exhaustive search is intractable because the number of all possible combinations
grows exponentially as $\prod_{i=1}^d t_i$.
Instead, we use a min-heap to efficiently keep track of the top $k$ combinations.
A min-heap is a complete binary tree, where each node is no greater than its children.
The operations of inserting an element and removing the smallest element from a min-heap can be done in logarithmic time.
\Cref{alg:maxksum} gives a procedure to solve the above problem using min-heaps.

\alglanguage{pseudocode}
\begin{algorithm}[ht]
  \caption{\texttt{maxk\_sum}: Combinations with the $k$ largest sums}
  \label{alg:maxksum}
  \begin{algorithmic}[1] %
    \Input two-dimensional array $A$; number of top combinations $k$.
    \State Make the array nonpositive by replacing $\mathbf{a}_i$ with $\mathbf{a}_i - a_{i, 1} \mathbf{1}$ for $i = 1, \cdots, d$.
    \State Create a min-heap by adding the elements of $\mathbf{a}_1$, each considered a combination of length one:
    index $I_1$, key $a_{1,I_1}$.
    \State At stage $i = 2, \cdots, d$: create a new min-heap consisting of length-$i$ combinations
    by adding each element in $\mathbf{a}_i$ to each combination in the min-heap at the previous stage:
    index $[I_1, \cdots, I_i]$, key $\sum_{j=1}^i a_{j,I_j}$.
    The size of the min-heap at each stage is capped at $k$
    by popping the smallest sum from the min-heap when necessary.
    \Output combinations in the min-heap at stage $d$.
  \end{algorithmic}
\end{algorithm}

This algorithm has time complexity $\mathcal{O}(t k \log k)$, where $t = \sum_{i=1}^d t_i \ll \prod_{i=1}^d t_i$,
and space complexity $\mathcal{O}(d k)$.
In TS-roots, the cost of \texttt{maxk\_sum} is small compared with the gradient-based multistart optimization of the posterior sample.

\section{Algorithms for TS-roots}
\label{apd:algorithms}

\subsection{Spectral Sampling of Separable Gaussian Process Priors}
\label{sub:spectral-sampling}

Per Mercer's theorem on probability spaces (see e.g., \cite{Rasmussen2006}, Sec 4.3),
any positive definite covariance function that is essentially bounded
with respect to some probability measure $\mu$ on a compact domain $\mathcal{X}$
has a spectral representation $\kappa(\mathbf{x}, \mathbf{x}') = \sum_{k=0}^\infty \lambda_k \phi_k(\mathbf{x}) \phi_k(\mathbf{x}')$,
where $(\lambda_k, \phi_k(\mathbf{x}))$ is a pair of eigenvalue and eigenfunction of the kernel integral operator.
The corresponding GP prior sample can be written as
$\fprior(\mathbf{x}) = \sum_{k=0}^{\infty} w_k \sqrt{\lambda_k} \phi_k(\mathbf{x})$,
where $w_k \iid \mathcal{N}(0,1)$ are independent standard Gaussian random variables.
Similar spectral representations exist per Bochner's theorem, which may have efficient discretizations \citep{Solin2020,Mutny2018}.

Given spectral representations of the univariate covariance functions of a separable Gaussian Process prior,
we can accurately approximate the prior sample as:
\begin{equation} \label{eq:spectral-sample-separable-prior}
    \fprior(\mathbf{x}) \approx \prod_{i=1}^{d} f_i(x_i; \randprior_i), \quad
    f_i(x_i; \randprior_i) \approx  \sum_{k=0}^{N_i-1} w_{i,k} \sqrt{\lambda_{i,k}} \phi_{i,k}(x_i).
\end{equation}
Here $N_i$ is selected for each variable such that $\lambda_{i,N_i-1}/\lambda_{i,1} \leq \eta_i$,
where $\eta_i$ is sufficiently small (see \Cref{apd:ExperimentDetails} for the value used in this study).
The product of the univariate sample functions forms a \textit{third-order approximation} of the multivariate sample function;
their moment functions differ at order four and higher even orders.
Using spectral representations of the univariate components as in \cref{eq:spectral-sample-separable-prior}
is much more efficient than directly using a spectral representation of the separable GP prior,
because the former uses $\sum_{i=1}^d N_i$ univariate terms to
approximate $\prod_{i=1}^d N_i$ multivariate terms in the latter.
Another benefit of this representation is that the sample functions are separable, which can be exploited in downstream tasks.

\paragraph{Spectrum of the Squared Exponential Covariance Function.}
\label{Specctrum}
The univariate squared exponential (SE) covariance function can be written as
$\kappa(x, x'; l) = \exp(-\frac{1}{2} s^2)$, where the relative distance $s = |x - x'| / l$ and length scale $l \in (0, \infty)$.
The spectral representation of such a covariance function per Mercer's theorem is $\kappa(x, x') = \sum_{k=0}^{\infty} \lambda_k \phi_k(x) \phi_k(x')$. 
With a Gaussian measure $\mu = \mathcal{N}(0, \sigma^2)$ over the domain $\mathcal{X} = \mathbb{R}$,
we can write the eigenvalues $\lambda_k$ and eigenfunctions $\phi_k(x)$ of the kernel integral operator as follows.
(See e.g., \cite{ZhuHY1998} Sec. 4 and \cite{Gradshteyn2014} 7.374 eq. 8.)

Define constants $a = (2 \sigma^{2})^{-1}$, $b = (2 l)^{-1}$, $c = \sqrt{a^2 + 4 a b}$, and
$A = \frac{1}{2} a + b + \frac{1}{2} c$.
For $k \in \mathbb{N}$, the $k$th eigenvalue is $\lambda_k = \sqrt{\frac{a}{A}} \left( \frac{b}{A}\right)^k$ and the corresponding eigenfunction is
$\phi_k(x) = \left( \frac{\pi c}{a} \right)^{1/4} \psi_k (\sqrt{c} x) \exp\left( \frac{1}{2} a x^2 \right)$,
where $\psi_k (x) = \left( \pi^{1/2} 2^k k! \right)^{-1/2} H_k(x) \exp\left( -\frac{1}{2} x^2 \right)$ and $H_k(x)$ the $k$th-order Hermite polynomial defined by $H_k(x) = (-1)^k \exp(x^2) \frac{d^k}{dx^k} \exp(-x^2)$.

\begin{figure}[t]
\centering
\includegraphics[scale = 0.9]{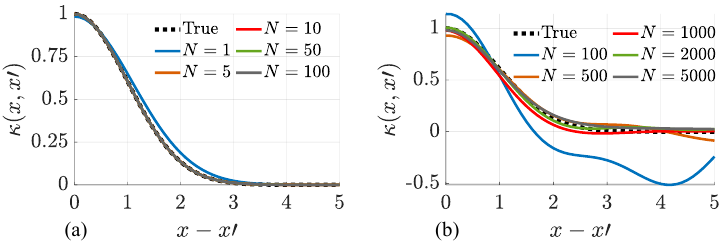}
\caption{Approximate SE covariance functions from
(a) the spectral representation per Mercer's theorem with the first $N$ eigenpairs and
(b) the random Fourier features representation with $N$ basis functions.
The plots are generated for $l=1$.} 
\label{fig:ApproxSE} 
\end{figure}

\Cref{fig:ApproxSE} shows approximations to the SE covariance function
by truncated spectral representations with the first $N$ eigenpairs
and by random Fourier features \citep{Rahimi2007} with $N$ basis functions.
The spectral representation per Mercer's theorem converges quickly to the true covariance function,
while the random Fourier features representation requires a large number of basis functions and is inaccurate for $N < 1000$.

\subsection{Univariate Global Rootfinding}

\Cref{alg:roots} outlines a method for univariate global rootfinding on an interval by solving an eigenvalue problem.
When the orthogonal polynomial basis is the Chebyshev polynomials, the corresponding comrade matrix is called a colleague matrix,
and we have the following theorem:

\alglanguage{pseudocode}
\begin{algorithm}[h]
	\caption{\texttt{roots}: Univariate global rootfinding on an interval}
	\label{alg:roots}
	\begin{algorithmic}[1] %
		\Input polynomial $p(x)$ of degree $m$ (or any real function $f(x)$)
		\State transform $p(x)$ into an orthogonal polynomial basis
		$p(x) = \sum_{k=0}^m a_k T_k(x)$\newline
		(or approximate $f(x)$ on the interval using such a basis)
		\State solve all the eigenvalues of the comrade matrix $\mathbf{C}$ associated with the polynomial basis
		\Output all the real eigenvalues $\{x_i\}_{i=1}^r$ in the interval, which are the roots of $p(x)$
		(or $f(x)$)
	\end{algorithmic}
\end{algorithm}

\begin{theorem} \label{theorem1}
Let $p(x) = \sum_{k=0}^m a_k T_k(x)$, $a_m \neq 0$, be a polynomial of degree $m$,
where $T_k$ is the $k$th Chebyshev polynomial and $a_k$ is the corresponding weight.
The roots of $p(x)$ are the eigenvalues of the following $m \times m$ colleague matrix:
\begin{equation} \label{eqn1}
	C=\left( 
	\begin{matrix}
		0 & 1& &  &  &  \\
		1/2&  0&  1/2&  &  &  \\
		&  1/2&  0&  1/2&  &  \\
		&  &  \ddots &  \ddots &  \ddots&  \\
		&  &  &  &  &  1/2\\
		&  &  &  &  1/2& 0
	\end{matrix}  \right) 
	-\frac{1}{2a_m} 
	\left( \begin{matrix}
		&  &  &  &  &  \\
		&  &  &  &  &  \\
		&  &  &  &  &  \\
		&  &  &  &  &  \\
		&  &  &  &  &  \\
		a_0&  a_1&  a_2&  \cdots&  \cdots& a_{m-1}
	\end{matrix} \right),
\end{equation}
where the elements not displayed are zero.
\end{theorem}

\clearpage
\alglanguage{pseudocode}
\begin{algorithm}[t!]
  \caption{\texttt{minsort}: Best local minima of a separable function}
  \label{alg:minsort}
  \begin{algorithmic}[1] %
    \Input
    separable function $f(\mathbf{x}) = \prod_{i=1}^d f_i(x_i)$;
    set size $\ncandidate$; buffer coefficient $\alpha$ (defaults to $3$).

    \State %
        $f_i(x_i) \gets \texttt{chebfun}(f_i(x_i))$, $i=1, \cdots, d$
    \Comment{Construct \texttt{chebfun}s for univariate components} \newline
    $f'_i(x_i) \gets \texttt{diff}(f_i(x_i)); \quad f''_i(x_i) \gets \texttt{diff}(f'_i(x_i))$
    \Comment{Compute first and second derivatives}

    \State $\{\xi_{i,j}\}_{j=1}^{r_i} \gets \texttt{roots}\left( f'_i (x_i) \right)$, $i=1, \cdots,d$
    \Comment{Univariate global rootfinding} \newline
    $\{\xi_{i,0}\} \gets \underline{x}_i; \quad \{\xi_{i,r_i + 1}\} \gets \overline{x}_i$
    \Comment{Include interval lower and upper bounds} \newline
    $\boldsymbol{\xi}_i \gets [\xi_{i,0},\xi_{i,1}, \cdots,\xi_{i,r_i},\xi_{i,r_i + 1}]^\intercal$
    \Comment{Candidate coordinate values $\{\xicandidate\}$}
    
    \State $\mathbf{f}_i \gets f_i(\boldsymbol{\xi}_i)$, $i=1, \cdots,d$
    \Comment{Univariate function values} \label{alg:minsort-line2} \newline
    $h_{i,j} \gets f''_i(\xi_{i,j})$, $j = 1, \cdots, r_i$
    \Comment{Univariate second derivatives at critical points} \newline
    $h_{i,0} \gets f'_i(\xi_{i,0})$; \quad $h_{i,r_i + 1} \gets -f'_i(\xi_{i,r_i + 1})$
    \Comment{Univariate inward derivatives at interval ends}
    
    \State $J_i \gets (\mathbf{f}_i \circ \mathbf{h}_i > 0); \quad P_i \gets (\mathbf{f}_i > 0)$
    \Comment{Boolean vectors of sign parity and positivity} \label{alg:minsort-line3}
    \State $\boldsymbol{\xi}_i^{(0)} \gets \boldsymbol{\xi}_i(J_i); \quad \boldsymbol{\xi}_i^{(1)} \gets \boldsymbol{\xi}_i(\neg J_i)$
    \Comment{Mono and mixed type candidate coordinates: $\xicandidatemono, \xicandidatemixed$} \newline
    $\mathbf{f}_i^{(0)} \gets \mathbf{f}_i(J_i); \quad \mathbf{f}_i^{(1)} \gets \mathbf{f}_i(\neg J_i)$
    \Comment{Values at mono and mixed type candidate coordinates}
    \State $n_i^{(0)} \gets \texttt{sum}(J_i); \quad n_i^{(1)} \gets \texttt{sum}(\neg J_i)$ \newline
    $n_i^{+(0)} \gets \texttt{sum}(P_i \texttt{\&} J_i); \quad n_i^{-(0)} \gets \texttt{sum}((\neg P_i) \texttt{\&} J_i)$ \newline
    $n_i^{+(1)} \gets \texttt{sum}(P_i \texttt{\&} (\neg J_i)); \quad n_i^{-(1)} \gets \texttt{sum}((\neg P_i) \texttt{\&} (\neg J_i))$ \newline
    $N^{(0)} \gets \prod_{i=1}^d n_i^{(0)}; \quad N^{(1)} \gets \prod_{i=1}^d n_i^{(1)}$
    \Comment{Sizes of tensor grids} \newline
    $S^{(0)} \gets \prod_{i=1}^d (n_i^{+(0)} - n_i^{-(0)}); \quad S^{(1)} \gets \prod_{i=1}^d (n_i^{+(1)} - n_i^{-(1)})$
    \Comment{Signed sums}
    
    \If{ $\ncandidate \le \breve{N}^{-} = \frac{1}{2} (N^{(1)} - S^{(1)})$ } \label{alg:minsort-line4}
    \State $\left[ \mathbf{s},\mathbf{I} \right] \gets \texttt{maxk\_sum} \left(\{ \log(|\mathbf{f}_i^{(1)}|) \}_{i=1}^d , \alpha \ncandidate\right)$
    \Comment{The $\alpha \ncandidate$ largest $|f|$ in $\Xmixed$} \label{alg:minsort-line14}
    \State $\mathbf{I} \gets \mathbf{I}[f(\boldsymbol{\xi}^{(1)}(\mathbf{I})) < 0, :]$
    \Comment{Multi-indices of best negative local minima} \label{alg:minsort-line-bnlm}
    \State $[\mathbf{b}, I] \gets \texttt{mink}(f(\boldsymbol{\xi}^{(1)}(\mathbf{I})), \ncandidate)$
    \Comment{The $\ncandidate$ smallest $f$ in $\Xneglocmin$} \label{alg:minsort-line-onlm}
    \State $\Xcandidate \gets \Xcandidate^{-} = \boldsymbol{\xi}^{(1)}(\mathbf{I}[I, :])$ \label{alg:minsort-line16}
        
    \Else
    \State $\boldsymbol{\Pi}^{(1)} \gets \prod_{i=1}^{d} \{1, \cdots, n_i^{(1)} \}$
    \Comment{Matrix of index combinations}
    \State $\breve{\mathbf{I}}^{-} \gets \boldsymbol{\Pi}^{(1)}[f(\boldsymbol{\xi}^{(1)}(\boldsymbol{\Pi}^{(1)})) < 0, :]$ 
    \Comment{Multi-indices of negative local minima} \label{alg:minsort-line-nlm}
    \State $[\mathbf{b}, I] \gets \texttt{sort}(f(\boldsymbol{\xi}^{(1)}(\breve{\mathbf{I}}^{-})))$
    \Comment{Sort values in ascending order}
    \State $\Xneglocmin \gets \boldsymbol{\xi}^{(1)}(\breve{\mathbf{I}}^{-}[I, :])$
    \Comment{Negative local minima}
    
    \If{ $\ncandidate \le \breve{N} = \frac{1}{2} (N^{(1)} - S^{(1)} + N^{(0)} + S^{(0)})$ }
    \State $\left[ \mathbf{s}, \mathbf{I} \right] \gets \texttt{maxk\_sum} \left(\{ \log(|\mathbf{f}_i^{(0)}|) \}_{i=1}^d , \alpha (\ncandidate - \breve{N}^{-}) \right)$
    \Comment{Largest $|f|$ in $\Xmono$} \label{alg:minsort-line14}
    \State $\mathbf{I} \gets \mathbf{I}[f(\boldsymbol{\xi}^{(0)}(\mathbf{I})) > 0, :]$
    \Comment{Multi-indices of best positive local minima} \label{alg:minsort-line-bplm}
    \State $[\mathbf{b}, I] \gets \texttt{mink}(f(\boldsymbol{\xi}^{(0)}(\mathbf{I})), \ncandidate - \breve{N}^{-})$
    \Comment{The $\ncandidate - \breve{N}^{-}$ smallest $f$ in $\Xposlocmin$} \label{alg:minsort-line-oplm}
    \State $\Xcandidate \gets \Xneglocmin \bigcup \Xcandidate^{+}, \; \Xcandidate^{+} = \boldsymbol{\xi}^{(0)}(\mathbf{I}[I, :])$ \label{alg:minsort-line16}

    \Else
    \State $\boldsymbol{\Pi}^{(0)} \gets \prod_{i=1}^{d} \{1, \cdots, n_i^{(0)} \}$
    \Comment{Matrix of index combinations}
    \State $\breve{\mathbf{I}}^{+} \gets \boldsymbol{\Pi}^{(0)}[f(\boldsymbol{\xi}^{(0)}(\boldsymbol{\Pi}^{(0)})) > 0, :]$
    \Comment{Multi-indices of positive local minima} \label{alg:minsort-line-plm}
    \State $[\mathbf{b}, I] \gets \texttt{sort}(f(\boldsymbol{\xi}^{(0)}(\breve{\mathbf{I}}^{+})))$
    \Comment{Sort values in ascending order}
    \State $\Xcandidate \gets \Xneglocmin \bigcup \Xposlocmin, \; \Xposlocmin = \boldsymbol{\xi}^{(0)}(\breve{\mathbf{I}}^{+}[I, :])$
    \Comment{All local minima} \label{alg:minsort-line9}
    \EndIf
    \EndIf
    
    \Output $\Xcandidate$
    \Comment{Candidate exploration set: smallest $\ncandidate$ local minima in ascending order}
  \end{algorithmic}
\end{algorithm}

\clearpage
A proof of \Cref{theorem1} is provided in \cite{Trefethen2019}, Chapter~18.
A classical formula to compute the weights $\{a_k\}$ requires $\bigO{(m^2)}$ floating point operations,
which can be reduced to $\bigO{(m \log m)}$ using a fast Fourier transform.
Since the colleague matrix is tridiagonal except in the final row,
the complexity of computing its eigenvalues can be improved from $\bigO{(m^3)}$ to $\bigO{(m^2)}$ operations,
which can be further improved to $\bigO{(m)}$ via recursive subdivision of intervals (see \citet{Trefethen2019}).
Specifically, if $m > 100$, the interval is divided recursively
so that on each subinterval the function can be accurately approximated by a polynomial of degree no greater than 100.
The \texttt{roots} algorithm is implemented in the Chebfun package in MATLAB \citep{Battles2004}
and the chebpy package in Python \citep{Richardson2016};
both packages also implement other related programs such as \texttt{chebfun} for Chebyshev polynomial approximation
and \texttt{diff} for differentiation.

\subsection{Best Local Minima of a Separable Function}

Given the univariate component functions of a separable function,
\Cref{alg:minsort} finds the subset $\Xcandidate$ of the local minima of the function with the $n_o$ smallest function values.
This procedure requires the \texttt{maxk\_sum} algorithm in \Cref{alg:maxksum},
the \texttt{roots} algorithm in \Cref{alg:roots} and the related programs \texttt{chebfun} and \texttt{diff},
see also \Cref{apd:ExperimentDetails}.

In \Cref{alg:minsort}, $\boldsymbol{\xi}, \mathbf{f}, \mathbf{h}, J, P$ are two-dimensional arrays,
while $\mathbf{I}, \boldsymbol{\Pi}^{(1)}, \boldsymbol{\Pi}^{(0)}$ are matrices.
Function evaluations at
\Cref{alg:minsort-line-bnlm,alg:minsort-line-onlm,alg:minsort-line-nlm,alg:minsort-line-bplm,alg:minsort-line-oplm,alg:minsort-line-plm}
are only notational: the sign and value of the function can be computed efficiently
by multiplying the signs and values of its components at the selected coordinates.
For example, the statement $f(\boldsymbol{\xi}^{(1)}(\mathbf{I})) < 0$ at \Cref{alg:minsort-line-bnlm} can be evaluated as
$\texttt{rowXor}(P^{(1)}(\mathbf{I}))$,
where $P^{(1)}$ is a two-dimensional array with $P_i^{(1)} = P_i(\neg J_i)$,
$P^{(1)}(\mathbf{I})$ is a matrix with $d$ columns,
and $\texttt{rowXor}$ is row-wise exclusive or operation.
Similarly, the statement $f(\boldsymbol{\xi}^{(1)}(\mathbf{I}))$ at \Cref{alg:minsort-line-onlm} can be evaluated as
$\texttt{rowProd}(\mathbf{f}^{(1)}(\mathbf{I}))$, where $\texttt{rowProd}$ is row products.

\subsection{Decoupled Sampling from Gaussian Process Posteriors}

The decoupled sampling method for GP posteriors \citep{Wilson2020},
together with the spectral sampling of separable GP priors, is outlined in \Cref{alg:samposteriorGP}.

\alglanguage{pseudocode}
\begin{algorithm}[b]
  \caption{Decoupled sampling of Gaussian process posterior}
  \label{alg:samposteriorGP}
  \begin{algorithmic}[1] %
    \Input eigenpairs $\{(\lambda_{i,k}, \phi_{i,k}(x))\}_{i = 1, \cdots, d}^{k = 0, \cdots, N_i-1}$,
    data $\mathcal{D} = \{ (\mathbf{x}^j, y^j) \}_{j=1}^n$,
    covariance matrix $\mathbf{C} = \mathbf{K}_{n,n} + \boldsymbol{\Sigma}$,
    canonical basis $\boldsymbol{\kappa}_{\cdot,n}(\mathbf{x}) = (\kappa(\mathbf{x}, \mathbf{x}^j))_{j=1}^n$.
    \State $w_{i,k} \iid \mathcal{N}(0,1)$ %
    \Comment{Random coefficients for the prior sample}
    \State $\fprior(\mathbf{x}) = \prod_{i=1}^{d} \sum_{k=0}^{N_i-1} w_{i,k} \sqrt{\lambda_{i,k}}  \phi_{i,k}(x_i)$
    \Comment{Approximate prior sample}
    \State $\mathbf{f}_n \gets [\fprior(\mathbf{x}^1), \cdots, \fprior(\mathbf{x}^n)]^\intercal$
    \Comment{Values of prior sample at observed locations}
    \State $\boldsymbol{\varepsilon} \sim \mathcal{N}_n(0, \boldsymbol{\Sigma})$ %
    \Comment{Random noise for the posterior sample}
    \State $\mathbf{v} \gets \mathbf{C}^{-1} \left( \mathbf{y} - \mathbf{f}_n - \boldsymbol{\varepsilon} \right)$
    \Comment{Linear solve via factorization (e.g., Cholesky or SVD)}
    \Output $\fpost(\mathbf{x}) = \fprior(\mathbf{x}) + \mathbf{v}^\intercal \boldsymbol{\kappa}_{\cdot,n}(\mathbf{x})$
    \Comment{Approximate posterior sample}
  \end{algorithmic}
\end{algorithm}

\subsection{Computational Complexity of TS-roots}
\label{sub:complexity}

Per \Cref{alg:TSroots}, the computational cost of the \texttt{TS-roots} method is dominated by a few tasks:
(1) one call of \texttt{minsort} (\Cref{alg:minsort});
(2) $\ncandidate + n$ evaluations of the posterior sample path $\widetilde{f}(\cdot)$;
and (3) $\nexplore + \nexploit$ calls of the gradient-based optimizer \texttt{minimize}.

First, consider task (2).
Evaluating $\widetilde{f}(\cdot)$ involves evaluating:
(i) the prior sample path $f(\cdot)$, which involves evaluating its $d$ univariate component functions,
each with a cost that depends on its spectral representation (\Cref{sub:spectral-sampling});
and (ii) the canonical basis $\boldsymbol{\kappa}_{\cdot,n}(\cdot)$, which costs $\mathcal{O}(d n)$ flops.
When the data size $n$ is large, we can pre-filter the observed locations $X$ by the observations $\mathbf{y}$,
which is a good estimate of $\widetilde{f}(X)$ depending on the observation noise.
We assume that the number of observed locations after filtering is at most comparable to $\nexplore$.
The cost of task (2) is thus $\mathcal{O}(\ncandidate d n)$ flops.

Now consider task (3).
Evaluating the gradient of $\widetilde{f}(\cdot)$ involves
evaluating the gradients of $f(\cdot)$ and $\boldsymbol{\kappa}_{\cdot,n}(\cdot)$.
Since both $f(\cdot)$ and $\boldsymbol{\kappa}_{\cdot,n}(\cdot)$ are separable functions,
their gradients can be computed at a cost comparable to evaluating their function values.
Let $N_{\text{grad}}$ be the number of gradient evaluations required by the gradient-based optimizer.
The cost of task (3) is thus $\mathcal{O}((\nexplore + \nexploit) N_{\text{grad}} d n)$ flops.

For task (1), the cost of the \texttt{minsort} algorithm is dominated by:
(i) $d$ calls to \texttt{chebfun}, which evaluates the univariate components $f_i(\cdot)$
at a number of points depending on their complexity;
(ii) $d$ calls to \texttt{roots} (\Cref{alg:roots}), which scales linearly with the polynomial degree $m$ of the chebfun object,
itself dependent on the complexity of $f_i(\cdot)$;
and (iii) at most one call to \texttt{maxk\_sum} (\Cref{alg:maxksum}) at a cost of $\mathcal{O}(t \ncandidate \log \ncandidate)$,
where $t = \sum_{i=1}^d t_i$ and $t_i$ is two plus the number of critical points of $f_i(\cdot)$
which depends on the complexity of $f_i(\cdot)$.
The complexity of $f_i(\cdot)$, for the SE kernel for example (see \Cref{sub:spectral-sampling}),
can be quantified as the inverse length scale $\theta_i = 1 / l_i$.
We may define an average complexity as $\theta = \frac{1}{d} \sum_{i=1}^d \theta_i$.
The cost of task (1) is thus $\mathcal{O}(d \theta \ncandidate \log \ncandidate)$ flops.

As explained in \Cref{apd:Minsize}, we can set $\nexplore$ and $\nexploit$ to small values and $\ncandidate$ to a moderate value,
independent of $\widetilde{f}(\cdot)$ and thus independent of $d$, $n$ and $\theta$.
The overall cost of \texttt{TS-roots} thus scales as $\mathcal{O}(d n + d \theta)$, which is linear in the input dimension $d$.

\section{Minimum Size of Exploration and Exploitation Sets}
\label{apd:Minsize}

We conduct an empirical experiment to determine minimal values for $\nexplore$ and $\nexploit$ of TS-roots algorithm, which are the sizes of $\Xexplore$ and $\Xexploit$, respectively. From this experiment, we also recommend a value for $\ncandidate$, which is the size of $\Xcandidate$.
Recall that points in $\Xcandidate$ are sorted in ascending order of prior sample values,
while those in $\Xexplore$ and $\Xexploit$ are sorted in ascending order of posterior sample values.

Let $I_\text{e}$ and $I_\text{x}$ be the sets of indices of points in $\Xexplore$ and $\Xexploit$ that converge to the best local minimum of the posterior sample in each optimization iteration, respectively.
Let $I_\text{o}$ be the set of indices of points in $\Xcandidate$ associated with $\Xexplore$. 
Our hypothesis is that we have a high chance of finding a small index value in either $I_\text{e}$ or $I_\text{x}$.
If this hypothesis is confirmed, then we can set both $\nexplore$ and $\nexploit$ at very small values,
which significantly accelerate the inner-loop optimization.
To confirm our hypothesis, we employ the following two steps.
First, we set $\nexplore$ and $\nexploit$ at large values to mimic the effect of removing the set size limits,
ensuring accurate solution of the global optimization problem.
Then, we show that we have a high chance of finding a small index value
from $I_\text{e}$ and/or $I_\text{x}$ in each optimization iteration.

\begin{figure}[t]
\centering
\includegraphics[width=\textwidth]{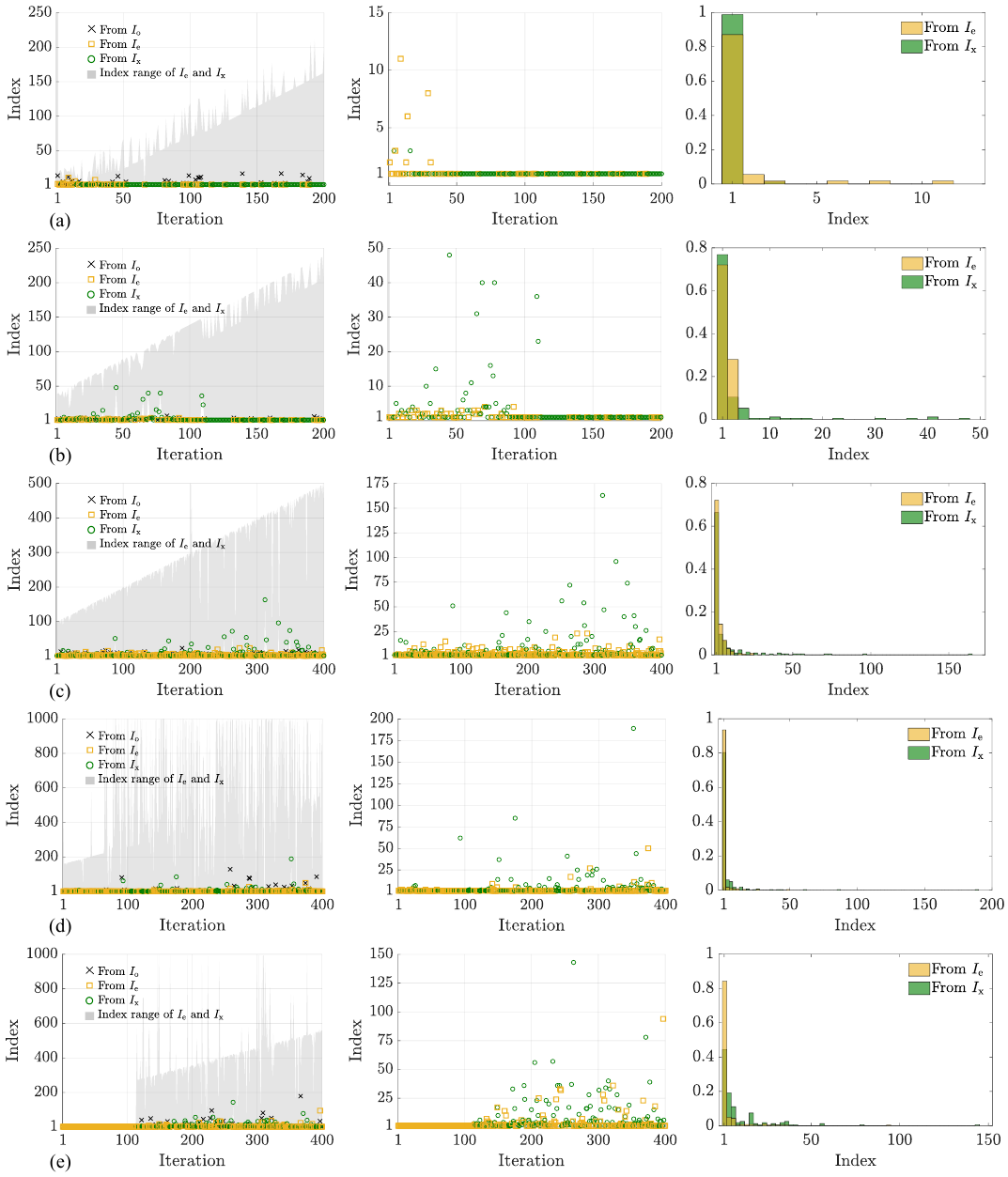}
\caption{\textit{Left column:} Minimum index values and index range of $I_\text{e}$ and/or $I_\text{x}$ for starting points that converge to the best local minimum $\bf{x}^\star$ of posterior sample in each optimization iteration, and index values of $I_\text{o}$ associated with minimum index values from $I_\text{e}$. \textit{Middle column:} Zoom-in plots of index values. \textit{Right column:} Historam of the minimum index values. (a) 2D Schwefel, (b) 4D Rosenbrock, (c) 10D Levy, (d) 16D Ackley, (e) 16D Powell functions.} 
\label{fig:index}
\end{figure}
\clearpage

We test our hypothesis on the 2D Schwefel, 4D Rosenbrock, 10D Levy, 16D Ackley, 16D Powell functions.
We set $\ncandidate = 5000$, $\nexplore = \nexploit = 1000$, and $\alpha=3$ (buffer coefficient).
The left and middle columns of \Cref{fig:index} show the smallest index values
and the variation of index values from $I_\text{e}$ and/or $I_\text{x}$ of starting points
that converge to the best local minimum $\bf{x}^\star$ of the posterior sample path in each optimization iteration.
The left column also plots the index values from $I_\text{o}$ corresponding to the smallest index values from $I_\text{e}$, if exists.
The right column shows the histograms of the smallest index values from $I_\text{e}$ and $I_\text{x}$ for all iterations considered.
These results show that we have a high chance of finding a small index value from $I_\text{e}$ and/or $I_\text{x}$ in each iteration.
This confirms our hypothesis.
In fact, using the first point in $\Xexplore$ and the first point in $\Xexplore$---only two points---we can
discover the global optimum most of the time.
Interestingly, the smallest index values appear largely independent of both the optimization iteration and the input dimension. Furthermore, the results suggest that it is safe to set $\ncandidate = 500$;
and for almost exact global optimization,
we suggest setting $\nexplore = 25$ and $\nexploit = 50$.

\alglanguage{pseudocode}
\begin{algorithm}[h!]
	\caption{Sequential optimization \citep{Garnett2023}}
	\label{alg:seqopt}
	\begin{algorithmic}[1] %
		\Input initial dataset $\mathcal{D}^0$
		\State $k \gets 1$
		\Repeat
		\State $\mathbf{x}^k \gets \mathrm{Policy}(\mathcal{D}^{k-1})$
		\State $y^k \gets \mathrm{Observe}(\mathbf{x}^k)$ 
		\State $\mathcal{D}^k \gets \mathcal{D}^{k-1} \cup \{(\mathbf{x}^k, y^k)\}$
		\Until{termination condition reached}
		\Output  $\mathcal{D}$
	\end{algorithmic}
\end{algorithm}

\alglanguage{pseudocode}
\begin{algorithm}[ht!]
	\caption{Bayesian optimization policy}
	\label{alg:BOpolicy}
	\begin{algorithmic}[1] %
		\Input a prior stochastic process $f$ for the objective function $\ftrue$, current dataset $\mathcal{D}^{k-1}$
		\State determine the posterior $f^k := f | \mathcal{D}^{k-1}$
		\State derive an acquisition function $\alpha^k(\mathbf{x})$ from $f^k$
		\State global optimization $\mathbf{x}^k \gets \argmin_{\mathbf{x} \in \mathcal{X}} \alpha^k(\mathbf{x})$
		\Output $\mathbf{x}^k$
	\end{algorithmic}
\end{algorithm}

\section{Bayesian Optimization via Thompson Sampling}
\label{apd:gp-ts}

A general procedure for sequential optimization is given in \Cref{alg:seqopt}.
The initial dataset $\mathcal{D}^0$ can either be empty or contain some observations.
In the latter case we can write $\mathcal{D}^0 = \{(\mathbf{x}^i, y^i)\}_{i=1}^{n_0}$, where $n_0 \in \mathbb{N}_{>0}$.
Three components of this algorithm can be customized:
the observation model $\mathrm{Observe}(\mathbf{x})$, the optimization policy $\mathrm{Policy}(\mathcal{D})$,
and the termination condition.

BO can be seen as an optimization policy for sequential optimization.
A formal procedure is given in \Cref{alg:BOpolicy}.
Three components of this algorithm can be customized: the prior probabilistic model $f$,
the acquisition function $\alpha$, and the global optimization algorithm.
Any probabilistic model of the objective function $\ftrue$ can be seen as a probability distribution on a function space,
and the prior $f$ is usually specified as a stochastic process such as a GP.
The acquisition function $\alpha$ derived from the posterior $f|\mathcal{D}$
can be either deterministic---such as EI and LCB---or stochastic, such as GP-TS.
To simplify notation, we state the global optimization problem of
$\alpha(\mathbf{x})$ as minimization rather than maximization.
The two problems are the same with a change of sign to the objective.

When applied to BO, GP-TS generates a random acquisition function simply by sampling the posterior model.
That is, given the posterior $f^k$ at the $k$th BO iteration,
the GP-TS acquisition function is a random function:
$\alpha^k(\mathbf{x}) \sim f^k$.

\section{Benchmark Functions}
\label{apd:BenchmarkFunctions}

The analytical expressions for the benchmark functions used in \Cref{results} are given below. The global solutions of these functions are detailed in \citep{Surjanovic2013}.

\paragraph{Schwefel Function:}
\begin{equation}
f(\mathbf{x}) = 418.9829 d - \sum_{i=1}^{d} x_i \sin \left( \sqrt{|x_i|} \right).
\end{equation}
This function is evaluated on $\mathcal{X}=[-500,500]^d$
and has a global minimum $f^\star := f(\mathbf{x}^\star) = 0$ at $\mathbf{x}^\star = [420.9687,\cdots,420.9687]^\intercal$.
This function is $C^1$ at $\mathbf{x} = 0$.

\paragraph{Rosenbrock Function:}
\begin{equation}
f(\mathbf{x}) = \sum_{i=1}^{d-1} \left[100(x_{i+1}-x_i^2)^2 + (x_i - 1)^2\right].
\end{equation}
This function is evaluated on $\mathcal{X}=[-5,10]^d$
and has a global minimum $f^\star = 0$ at $\mathbf{x}^\star = [1,\cdots,1]^\intercal$.

\paragraph{Levy Function:}
\begin{equation}
f(\mathbf{x}) = \sin^2 (\pi w_1) + \sum_{i=1}^{d-1} (w_i - 1)^2 \left[ 1 + 10 \sin^2 (\pi w_i +1) \right] + (w_d - 1)^2 \left[ 1 +  \sin^2 (2 \pi w_d) \right],
\end{equation}
where $w_i = 1 + \frac{x_i-1}{4}$, $i = 1,\cdots,d$.
This function is evaluated on $\mathcal{X}=[-10,10]^d$
and has a global minimum $f^\star = 0$ at $\mathbf{x}^\star = [1,\cdots,1]^\intercal$.

\paragraph{Ackley Function:}
\begin{equation}
f(\mathbf{x}) = -a \exp\left(-b\sqrt{\frac{1}{d} \sum_{i=1}^{d}x^2_i}\right) - \exp\left(\frac{1}{d} \sum_{i=1}^{d}\cos(cx_i)\right) +a + \exp(1),
\end{equation}
where $a = 20$, $b = 0.2$, and $c = 2\pi$.
This function is evaluated on $\mathcal{X}=[-10,10]^d$
and has a global minimum $f^\star = 0$ at $\mathbf{x}^\star = [0,\cdots,0]^\intercal$.
This function not differentiable at $\mathbf{x}^\star$.

\paragraph{Powell Function:}
\begin{equation}
f(\mathbf{x}) =  \sum_{i=1}^{d/4} \left[ \left(x_{4i-3} + 10x_{4i-2}\right)^2 + 5\left(x_{4i-1} - x_{4i}\right)^2 + \left(x_{4i-2} - 2x_{4i-1}\right)^4 + 10\left(x_{4i-3} - x_{4i}\right)^4 \right].
\end{equation}
This function is evaluated on $\mathcal{X}=[-4,5]^d$
and has a global minimum $f^\star = 0$ at $\mathbf{x}^\star = [0,\cdots,0]^\intercal$.

\paragraph{6d Hartmann Function:}
\begin{equation}
f(\mathbf{x}) = -\sum_{i=1}^{4} a_i \exp \left( -\sum_{j=1}^{6} A_{ij} (x_j-P_{ij})^2 \right),
\end{equation}
where 
\begin{subequations}
	\begin{equation}
		\mathbf{a} = [1,1.2,3,3.2]^\intercal,
	\end{equation}
	\begin{equation}
		\mathbf{A}  = 
		\begin{bmatrix}
			10 & 3 & 17 & 3.5 & 1.7 & 8 \\
			0.05 & 10 & 17 & 0.1 & 8 & 14 \\
			3 & 3.5 & 1.7 & 10 & 17 & 8 \\
			17 & 8 & 0.05 & 10 & 0.1 & 14
		\end{bmatrix},
	\end{equation}
	\begin{equation}
		\mathbf{P}  = 10^{-4}\begin{bmatrix}
			1312 &1696 &5569 &124 &8283 &5886\\
			2329 &4135 &8307 &3736 &1004 &9991\\
			2348 &1451 &3522 &2883 &3047 &6650\\
			4047 &8828 &8732 &5743 &1091 &381
		\end{bmatrix}
	\end{equation}
\end{subequations}
This function is evaluated on $\mathcal{X}=[0,1]^6$
and has a global minimum $f^\star = -3.32237$ at $\mathbf{x}^\star = [0.20169,0.150011,0.476874,0.275332,0.311625,0.6573]^\intercal$.
The rescaled version $\widetilde{f}(\mathbf{x}) = \frac{f(\mathbf{x}) - 2.58}{1.94}$ \citep{Picheny2013} is used in the experiments.

\begin{figure}[t]
	\centering
	\includegraphics[width=0.45\textwidth]{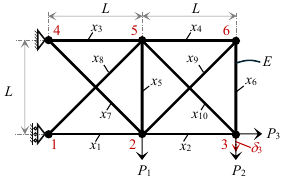}
	\caption{Ten-bar truss. Cross-sectional areas of ten truss members are the input variables $x_i$, $i \in \{1\dots,10$\}. Known parameters include length $L$, Young's modulus of truss material $E$, and external loads $P_j$, $j\in\{1,2,3\}$.
    The vertical displacement at node 3 is denoted as $\delta_3$.}
	\label{fig:10bartruss}
\end{figure}

\section{Ten-bar Truss}
\label{apd:10bartruss}

Consider a ten-bar truss shown in \Cref{fig:10bartruss}.
The truss has ten members and is subjected to vertical load $P_1 = 60$ kN at node 2, vertical load $P_2 = 40$ kN at node 3, and horizontal load $P_3 = 10$ kN at node 3. The Young's modulus of the truss material $E = 200$ GPa. The length parameter $L=1$ m. Let $A({\bf x})=\sum_{i=1}^{10}x_i$ and $\delta_3({\bf x})$ denote the total area of the cross-sectional areas of the truss members and the vertical displacement at node 3, respectively, where ${\bf x} = [x_1,\dots,x_{10}]^\intercal$ is the vector of cross-sectional areas of the truss members.
The optimization problem formulated for the truss is to minimize both $A({\bf x})$ and $\delta_3({\bf x})$. Since $A({\bf x})$ and $\delta_3({\bf x})$ are competing, we define the objective function as a weight-sum of $A({\bf x})$ and $\delta_3({\bf x})$, such that
\begin{equation}
    f({\bf x})=w_1\frac{A({\bf x})}{A_{\max}} + w_2\frac{\delta_3({\bf x})}{\delta_{\max}},
\end{equation}
where ${\bf x} \in [1,20]^{10} \, \mathrm{cm}^2$, $w_1 = 0.6$, $w_2 = 0.4$, $A_{\max}=200 \, \mathrm{cm}^2$, and $\delta_{\max} = 3 \, \mathrm{cm}$.

\section{Experimental Details}
\label{apd:ExperimentDetails}

\paragraph{Data Generation.}
We generate 20 initial datasets for each problem.
The input observations are randomly generated using the Latin hypercube sampling \citep{Owen1992} within $[-1,1]^d$,
where $d$ represents the number of input variables.
The normalized input observations are transformed into their real spaces to evaluate the corresponding objective function values
which are then standardized using the $z$-score for processing optimization.
Each BO method in comparison starts from each of the generated datasets.

\paragraph{Key Parameters for TS-roots and other BO Methods.}
We use squared exponential (SE) covariance functions for our experiments.
The spectra of univariate SE covariance functions for all problems (see \Cref{Specctrum})
are determined using the Gaussian measure $\mu = \mathcal{N}(0, 1)$.
The number of terms $N_i$, $i \in \{1,\cdots,d\}$, of each truncated univariate spectrum is determined such that
$\lambda_{i,N_i-1}/\lambda_{i,1} \leq \eta_i$, where $\eta_i = 10^{-16}$.
If $N_i > 1000$, we set $N_i = 1000$ to trade off between the accuracy of truncated spectra and computational cost.
We also set $\ncandidate = 500$.
The maximum size of the exploration set is $\nexplore = 250$. 
The maximum size of the exploitation set is $\nexploit = 200$.

The number of initial observations is $10d$ for all problems.
The standard deviation of observation noise $\sigma_\text{n} = 10^{-6}$ is applied for standardized output observations.
The number of BO iterations for the 2D Schwefel and 4D Rosenbrock functions is 200, while that for the 10D Levy, 16D Ackley, and 16D Powell functions is 800.
Other GP-TS methods for optimization of benchmark test functions including TS-DSRF (i.e., TS using decoupled sampling with random Fourier
features) and TS-RF (i.e., TS using random Fourier features) are characterized by a total of 2000 random Fourier features.

To ensure a fair comparison of outer-optimization results, we first implement TS-roots and record the number of starting points used in each optimization iteration. We then apply other BO methods, each employing a gradient-based multistart optimizer with the same number of random starting points and identical termination criteria as those used for TS-roots in each iteration.

For the comparative inner-loop optimization performance of the proposed method via rootfinding with the random multistart and genetic algorithm approaches, we set the same termination tolerance on the objective function value as the stopping criterion for the methods. In addition, the number of starting points for the random multistart and the population size of the genetic algorithm are the same as the number of points in both the exploration and exploitation sets of rootfinding in each optimization iteration.

\paragraph{Computational Tools.}

We carry out all experiments, except those for inner-loop optimization, using a designated cluster at our host institution. This cluster hosts 9984 Intel CPU cores and 327680 Nvidia GPU cores integrated within 188 compute and 
20 GPU nodes. 
The inner-loop optimization is implemented on a PC with an Intel\textsuperscript{\textregistered} Core\textsuperscript{TM} i7-1165G7 @ 2.80 GHz and 16 GB memory.

For the univariate global rootfinding via Chebyshev polynomials, we use MATLAB's Chebfun package \citep{Battles2004}
and its corresponding implementation in Python, called chebpy \citep{Richardson2016}.

\clearpage
\section{Additional Results}
\label{apd:AdditionalResults}

\begin{figure}[t]
	\centering
	\includegraphics[width=0.95\textwidth]{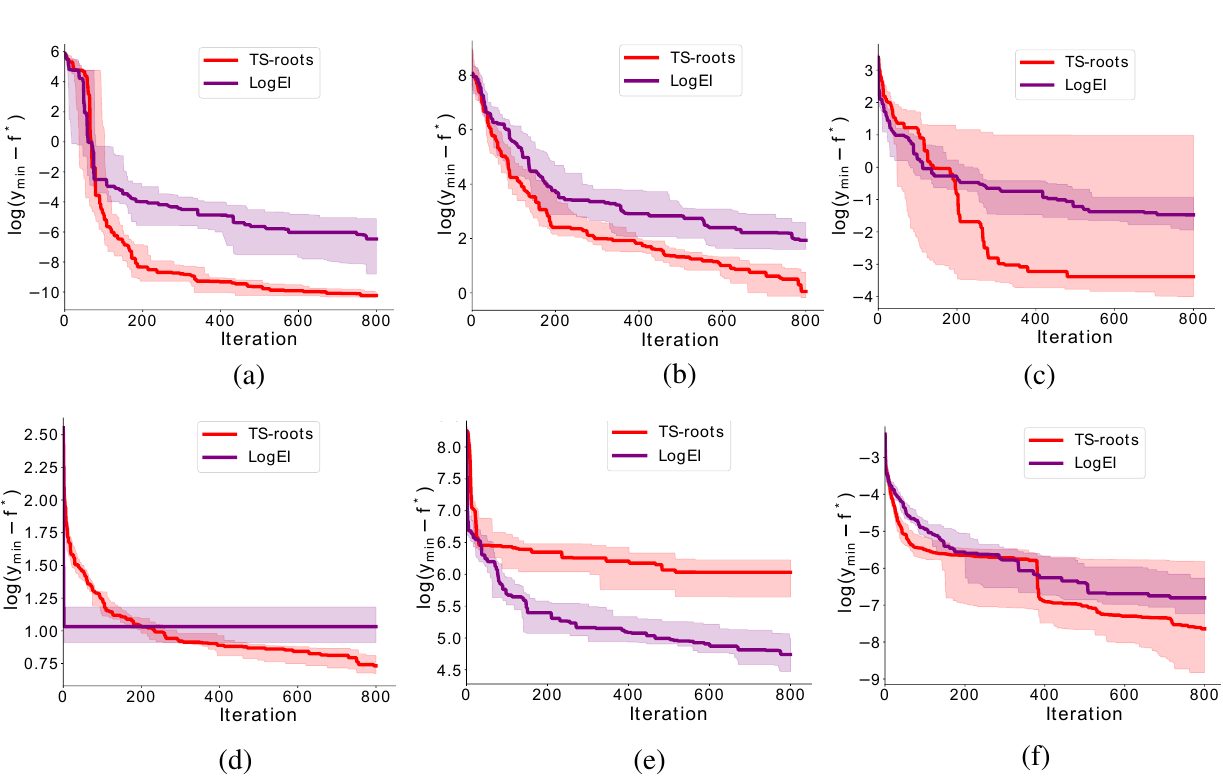}
	\caption{Outer-loop optimization results for the (a) 2D Schwefel function, (b) 4D Rosenbrock function, (c) 10D Levy function, (d) 16D Ackley function, (e) 16D Powell function, and (f) ten-bar truss problem.
The plots are histories of medians and interquartile ranges of solution values from 20 runs of TS-roots and LogEI.}
	\label{fig:outerTSLogEI}
\end{figure}

\paragraph{Comparison of TS-roots and LogEI Outer-loop Optimization Results.}
\Cref{fig:outerTSLogEI} compares the performance of the outer-loop optimization of TS-roots and LogEI \citep{Ament2023} on the five benchmark functions and a real-world ten-bar truss structure. Across all examples, save 16D Powell, TS-roots outperforms LogEI, further demonstrating the robustness of our approach. The comparatively weaker performance on the 16D Powell function can be attributed to its convex landscape, where BO generally underperforms relative to gradient-based methods.

\paragraph{Distance to Global Minimum.}

\begin{figure}[tb]
	\centering
	\includegraphics[width=\textwidth]{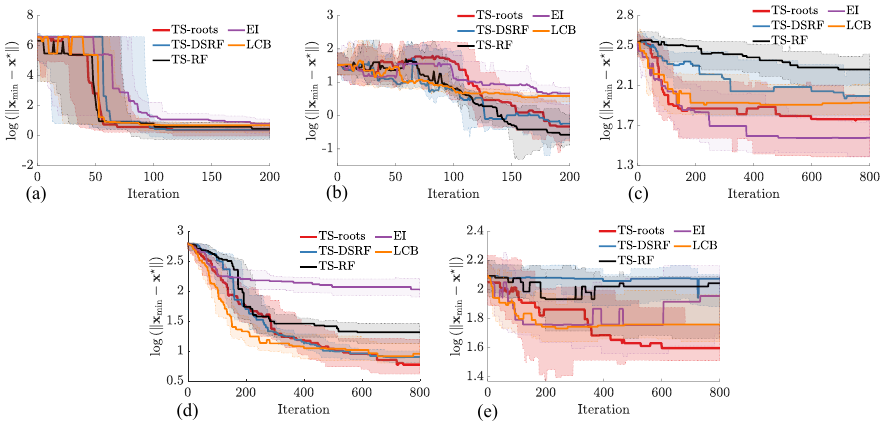}
	\caption{ Outer-loop optimization results for (a) the 2D Schwefel, (b) 4D Rosenbrock, (c) 10D Levy, (d) 16D Ackley, (e) 16D Powell functions. The plots are histories of medians and interquartile ranges of solution locations from 20 runs of TS-roots, TS-DSRF (i.e., TS using decoupled sampling with random Fourier features), TS-RF (i.e., TS using random Fourier features), EI, and LCB.}
	\label{fig:OptResultsLocations}
\end{figure}

\Cref{fig:OptResultsLocations} shows the solution locations from 20 runs of TS-roots, TS-DSRF, TS-RF, EI, and LCB for the 2D Schwefel, 4D Rosenbrock, 10D Levy, 16D Ackley, 16D Powell functions.

\paragraph{Comparison of Inner-loop Optimization Results.}

\begin{figure}[t]
	\centering
	\includegraphics[width=\textwidth]{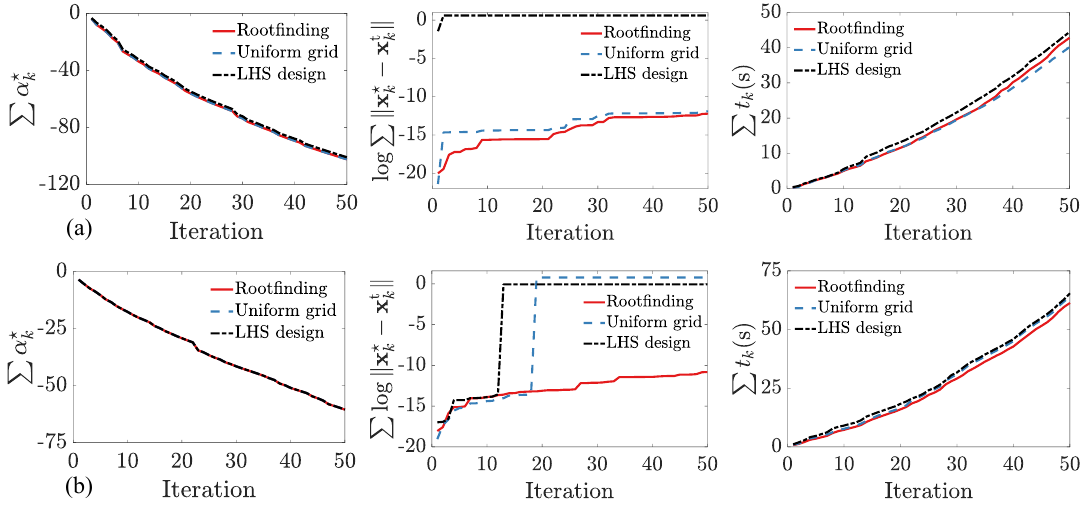}
	\caption{Inner-loop optimization results by three different initialization schemes, i.e., rootfinding, uniform grid, and Latin hypercube sampling, for (a) the 2D Schwefel and (b) 4D Rosenbrock functions. The plots are cumulative values of optimized GP-TS acquisition functions $\alpha_k^\star$, cumulative distances between new solution points $\mathbf{x}_k^\star$ and the true global minima $\mathbf{x}_k^\text{t}$ of the acquisition functions, and cumulative CPU times $t_k$ for optimizing the acquisition functions.}
	\label{fig:innercompare3}
\end{figure}

\begin{figure}[ht!]
	\centering
	\includegraphics[width=\textwidth]{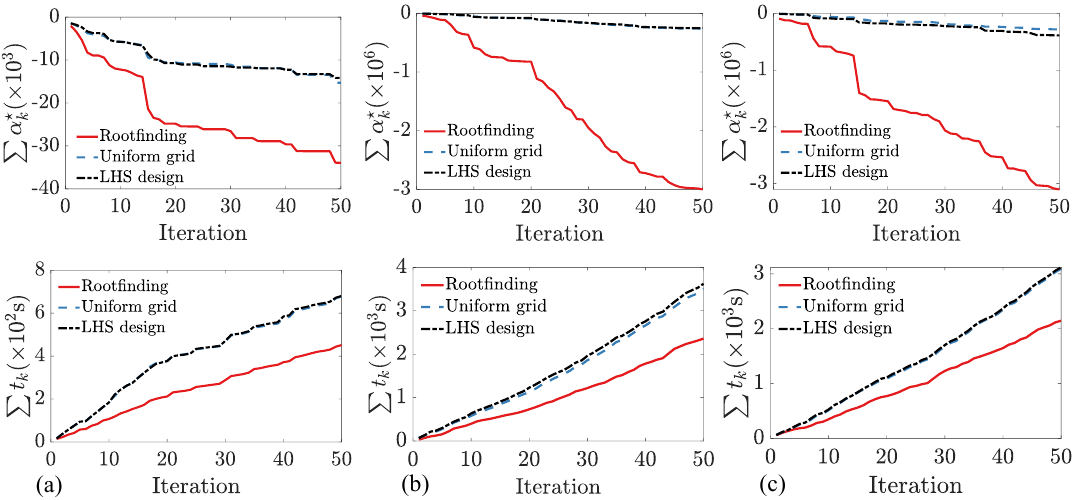}
	\caption{Inner-loop optimization results by three different initialization schemes, i.e., rootfinding, uniform grid, and Latin hypercube sampling, for (a) the 10D Levy, (b) 16D Ackley, and (c) 16D Powell functions. The plots are cumulative values of optimized GP-TS acquisition functions $\alpha_k^\star$ and cumulative CPU times $t_k$ for optimizing the acquisition functions.}
	\label{fig:innercompare4}
\end{figure}

\Cref{fig:innercompare3,fig:innercompare4} compare the performance of the inner-loop optimization by three different initialization schemes, i.e., rootfinding, uniform grid, and Latin hypercube sampling, for low-dimensional cases of the 2D Schwefel and 4D Rosenbrock functions, and for higher-dimensional cases of the 10D Levy, 16D Ackley, and 16D Powell functions. Rootfinding performs better than the uniform grid and Latin hypercube sampling initialization schemes, especially in high-dimensional settings.

\paragraph{Sample-average Posterior Function.}

\begin{figure}[t]
\centering
\includegraphics[width=0.45\textwidth]{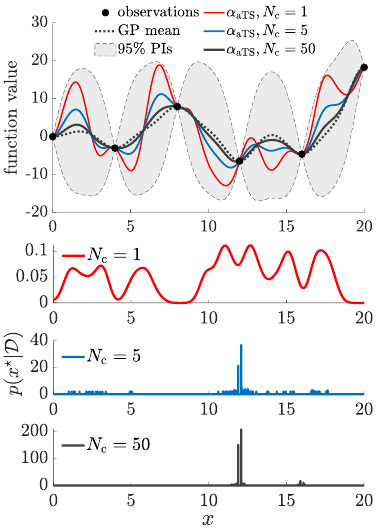}
\caption{Sample-average posterior function for different values of $N_\text{c}$. The posterior function approaches the GP mean and the conditional distribution of the solution location $p(x^\star|\mathcal{D})$ is more concentrated when we increase $N_\text{c}$.}
\label{fig:averTSConcept}
\end{figure}

\Cref{fig:averTSConcept} shows how we can improve the exploitation of GP-TS when increasing the exploration--exploitation control parameter $N_\text{c}$.

\paragraph{Performance of Sample-average TS-roots.}

\begin{figure}[ht!]
\centering
\includegraphics[width=\textwidth]{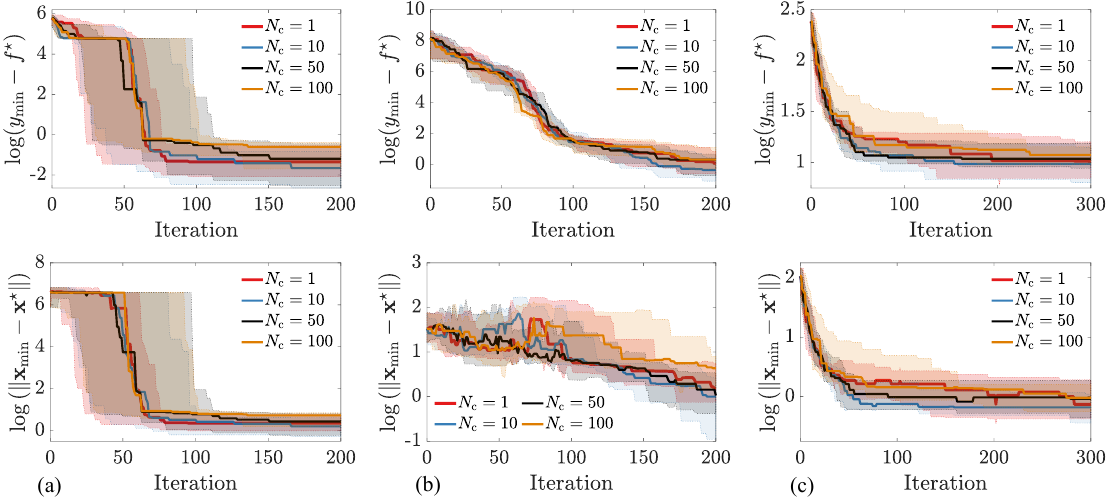}
\caption{ Performance of sample-average TS-roots with different control values $N_\text{c}$ for (a) the 2D Schwefel, 4D Rosenbrock, and (b) 6D Ackley functions. The plots are histories of medians and interquartile ranges of solution values and solution locations from 20 runs of TS-roots for each $N_\text{c}$ value.}
\label{fig:averTSResults}
\end{figure}

\Cref{fig:averTSResults} shows the performance of sample-average TS-roots with different exploration--exploitation control parameters $N_\text{c}$ for the 2D Schwefel, 4D Rosenbrock, and 6D Ackley functions.

\end{document}